\documentclass[runningheads]{llncs}
\usepackage{graphicx}
\usepackage{comment}
\usepackage{amsmath,amssymb} % define this before the line numbering.
\usepackage{color}

\usepackage[width=122mm,left=12mm,paperwidth=146mm,height=193mm,top=12mm,paperheight=217mm]{geometry}
\usepackage{subfiles}
\usepackage{xcolor}
\usepackage{multirow}
\usepackage{diagbox}
\usepackage{booktabs}
\newcommand{\ra}[1]{\renewcommand{\arraystretch}{#1}}
\usepackage[pagebackref=true,breaklinks=true,letterpaper=true,colorlinks,bookmarks=false]{hyperref}
\usepackage{microtype} 
\usepackage{ulem}
\usepackage{threeparttable}

\makeatletter
\newcommand{\printfnsymbol}[1]{%
  \textsuperscript{\@fnsymbol{#1}}%
}
\makeatother

\begin{document}
% \linenumbers
\pagestyle{headings}
\mainmatter
\def\ECCVSubNumber{1485}  % Insert your submission number here
\title{CelebA-Spoof: Large-Scale Face Anti-Spoofing Dataset with Rich Annotations} % Replace with your title

\begin{comment}
\titlerunning{ECCV-20 submission ID \ECCVSubNumber} 
\authorrunning{ECCV-20 submission ID \ECCVSubNumber} 
\author{Anonymous ECCV submission}
\institute{Paper ID \ECCVSubNumber}
\end{comment}
%******************

% \begin{comment}
\titlerunning{CelebA-Spoof}
%\orcidID{0000-0002-9063-7886}
\author{Yuanhan Zhang\inst{1} $^{,}$ \inst{2} \thanks{equal contribution} \and
Zhenfei Yin\inst{2} \printfnsymbol{1} \and 
Yidong Li\inst{1} \and \\ Guojun Yin \inst{2}\and  Junjie Yan \inst{2}\and Jing Shao \inst{2}\and Ziwei Liu \inst{3}}
\authorrunning{Zhang et al.}
\institute{Beijing Jiaotong University, Beijing, China \and
SenseTime Group Limited \and The Chinese University of Hong Kong \\
\email{\{18120454,yidongli\}@bjtu.edu.cn} 
\and
\email{\{yinzhenfei,yinguojun,yanjunjie,shaojing\}@sensetime.com} \and \email{zwliu@ie.cuhk.edu.hk}}
% \end{comment}
%******************
\maketitle

% \subfile{title.tex}

\begin{abstract}
% Face anti-spoofing is one of the key problems in computer vision. Despite the community's efforts in data collection, few existing face anti-spoofing datasets can cover a wide range of scenes, subjects and input sensors with dense and detailed annotation for face anti-spoofing. In this paper, we introduce and analyze the CelebA-Spoof Dataset with 625537 pictures of 10177 subjects, 8 scenes (2 environments * 4 illumination conditions) and build on 43 semantic attributes on CelebA-Spoof. Based on CelebA-Spoof, we explore the pros and cons of auxiliary geometric information which are widely used in face anti-spoofing algorithm. And extensive experiments show that using attributes as auxiliary semantic supervision display significant performance improvement in face anti-spoofing task compared to basic binary supervision.

As facial interaction systems are prevalently deployed, security and reliability of these systems become a critical issue, with substantial research efforts devoted. Among them, face anti-spoofing emerges as an important area, whose objective is to identify whether a presented face is live or spoof. Though promising progress has been achieved, existing works still have difficulty in handling complex spoof attacks and generalizing to real-world scenarios. The main reason is that current face anti-spoofing datasets are limited in both quantity and diversity. To overcome these obstacles, we contribute a large-scale face anti-spoofing dataset, \textbf{CelebA-Spoof}, with the following appealing properties: \textit{1) Quantity:} CelebA-Spoof comprises of 625,537 pictures of 10,177 subjects, significantly larger than the existing datasets. \textit{2) Diversity:} The spoof images are captured from 8 scenes (2 environments * 4 illumination conditions) with more than 10 sensors. \textit{3) Annotation Richness:} CelebA-Spoof contains 10 spoof type annotations, as well as the 40 attribute annotations inherited from the original CelebA dataset. Equipped with CelebA-Spoof, we carefully benchmark existing methods in a unified multi-task framework, \textbf{Auxiliary Information Embedding Network (AENet)}, and reveal several valuable observations. Our key insight is that, compared with the commonly-used binary supervision or mid-level geometric representations, rich semantic annotations as auxiliary tasks can greatly boost the performance and generalizability of face anti-spoofing across a wide range of spoof attacks. Through comprehensive studies, we show that CelebA-Spoof serves as an effective training data source. Models trained on CelebA-Spoof (without fine-tuning) exhibit state-of-the-art performance on standard benchmarks such as CASIA-MFSD. The 
dataset is available at: \href{https://github.com/Davidzhangyuanhan/CelebA-Spoof}{https://github.com/Davidzhangyuanhan/CelebA-Spoof}. 

\keywords{Face Anti-Spoofing, Large-Scale Dataset}

\end{abstract}

\section{Introduction}

Face anti-spoofing is an important task in computer vision, which aims to facilitate facial interaction systems to determine whether a presented face is live or spoof. With the successful deployments in phone unlock, access control and e-wallet payment, facial interaction systems already become an integral part in the real world. However, there exists a vital threat to these face interaction systems. Imagine a scenario where an attacker with a photo or video of you can unlock your phone and even pay his bill using your e-wallet. To this end, face anti-spoofing has emerged as a crucial technique to protect our privacy and property from being illegally used by others.

% In the past decade, many spoof attack threats appear, such as \textit{Print}, \textit{Replay}, \textit{3D Mask}, \textit{Paper Cut}, \textit{etc.} These attacks deceive the facial interaction systems and make these systems incorrectly identify the attacker as the genuine user. In response to such threats, most of the traditional methods focus on designing hand-crafted features~\cite{HoG:schwartz2011face,LBP:chingovska2012effectiveness,Surf_YCBCr:boulkenafet2016face} to detect the appearance change. Other works are interested in exploiting temporal information~\cite{eyeblink:pan2007eyeblink,lipmotion:kollreider2007real}. With the resurgence of deep learning, more recent solutions treat face anti-spoofing as a binary classification problem~\cite{CNN_based:feng2016integration,CNN_based:li2016original,STASN} and achieve promising performance. Existing methods have also explored auxiliary information (\textit{e.g.} mid-level geometric representations )~\cite{SiW,kim2019basn,depth:atoum2017face,deep_domain_generalization_2019_CVPR} to further improve the performance. 

% All the aforementioned methods are fueled by the availability of face anti-spoofing datasets~\cite{SiW,oulu-NPU,casiasurf,MSU-USSA,Replay-Attack,MSU-MFSD,CASIA-MFSD}, as shown in Table~\ref{tab:Dataset}. 
Most modern face anti-spoofing methods~\cite{CNN_based:feng2016integration,CNN_based:li2016original,STASN} are fueled by the availability of face anti-spoofing datasets~\cite{SiW,oulu-NPU,casiasurf,MSU-USSA,Replay-Attack,MSU-MFSD,CASIA-MFSD}, as shown in Table~\ref{tab:Dataset}. 
However, there are several limitations with the existing datasets: 1) \textit{Lack of Diversity.} Existing datasets suffer from lacking sufficient subjects, sessions and input sensors (\textit{e.g.} mostly less than 2000 subject, 4 sessions and 10 input sensors). 2) \textit{Lack of Annotations.}  Existing datasets have only annotated the type of spoof type. Face anti-spoof community lacks a densely annotated dataset covering rich attributes, which can further help researchers to explore face anti-spoofing task with diverse attributes. 3) \textit{Performance Saturation.} The classification performance on several face anti-spoofing datasets has already saturated, failing to evaluate the capability of existing and future algorithms. For example, the recall under FPR = 0.5\% on SiW and Oulu-NPU datasets using vanilla ResNet-18 has already reached 100.0\% and 99.0\%, respectively.

\begin{figure}[t]
\centering
\includegraphics[width=\textwidth]{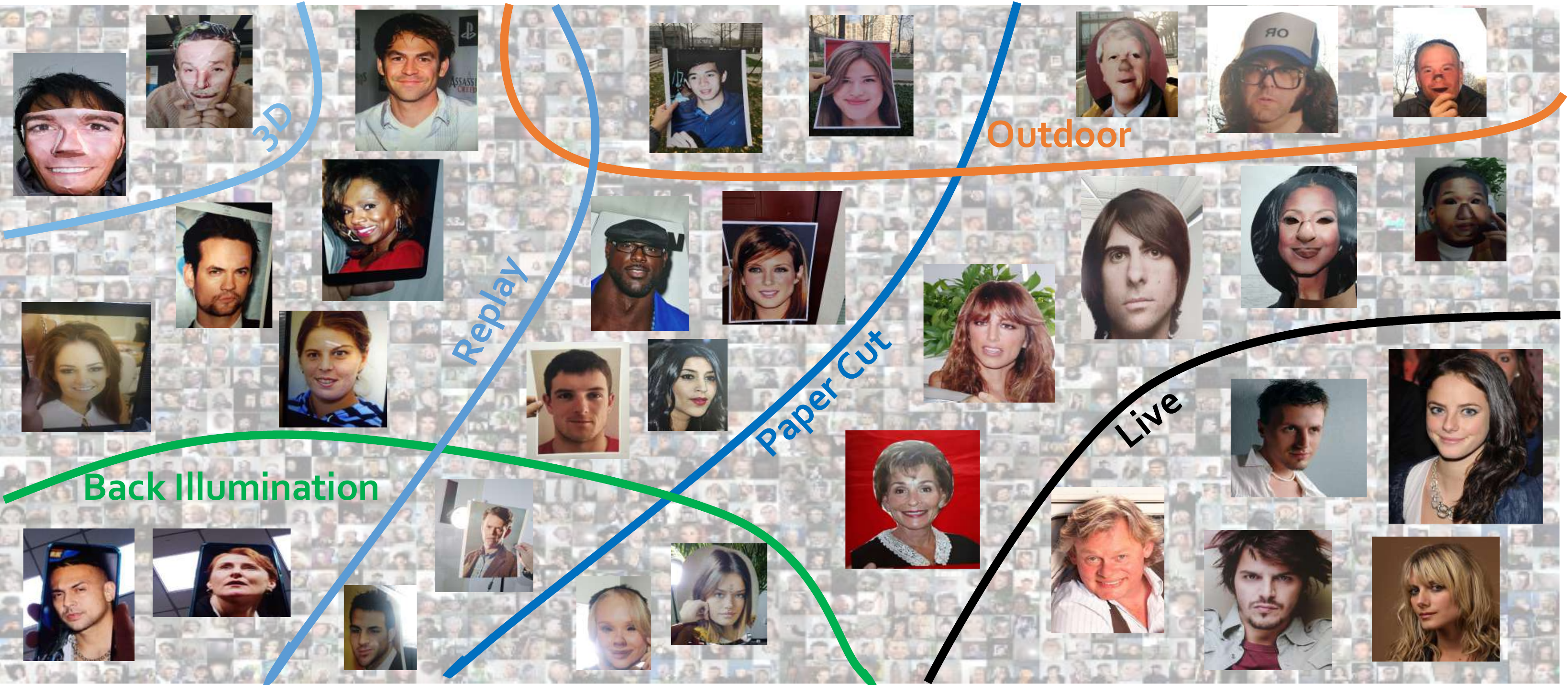}
\caption{A quick glance of CelebA-Spoof face anti-spoofing dataset with its attributes. Hypothetical space of scenes are partitioned by attributes and Live/Spoof. In reality, this space is much higher dimensional and there are no clean boundaries between attributes presence and absence}
\label{fig:fig1}
\end{figure}

To address these shortcomings in existing face anti-spoofing dataset, in this work we propose a large-scale and densely annotated dataset, \textbf{CelebA-Spoof}. Besides the standard \textit{Spoof Type} annotation, CelebA-Spoof also contains annotations for \textit{Illumination Condition} and \textit{Environment}, which express more information in face anti-spoofing, compared to categorical label like \textit{Live/Spoof}. Essentially, these dense annotations describe images by answering questions like ``Is the people in the image Live or Spoof?'', ``What kind of spoof type is this?'', ``What kind of illumination condition is this?'' and ``What kind of environment in the background?''. Specifically, all live images in CelebA-Spoof are selected from CelebA \cite{CelebA}, and all Spoof images are collected and annotated by skillful annotators. CelebA-Spoof has several appealing properties. \textbf{1) Large-Scale.} CelebA-Spoof comprises of a total of 10177 subjects, 625537 images, which is the largest dataset in face anti-spoofing. \textbf{2) Diversity.} For collecting images, we use more than 10 different input tensors, including phones, pads and personal computers (PC). Besides, we cover images in 8 different sessions. \textbf{3) Rich Annotations.} Each image in CelebA-Spoof is defined with 43 different attributes: 40 types of \textit{Face Attribute} defined in CelebA \cite{CelebA} plus 3 attributes of face anti-spoofing, including: \textit{Spoof Type}, \textit{Illumination Condition} and \textit{Environment}. With rich annotations, we can comprehensively investigate face anti-spoofing task from various perspectives.

Equipped with CelebA-Spoof, we design a simple yet powerful network named \textbf{A}uxiliary information \textbf{E}mbedding \textbf{N}etwork (\textbf{AENet}), and carefully benchmark existing methods within this unified multi-task framework. Several valuable observations are revealed: \textbf{1)} We analyze the effectiveness of auxiliary \textbf{geometric information} for different spoof types and illustrate the sensitivity of geometric information to special illumination conditions. Geometric information includes \textit{depth map} and \textit{reflection map}. \textbf{2)} We validate auxiliary \textbf{semantic information}, including face attribute and spoof type, plays an important role in improving classification performance. \textbf{3)} We build three CelebA-Spoof benchmarks based on this two auxiliary information. Through extensive experiments, we demonstrate that our large-scale and densely annotated dataset serves as an effective data source in face anti-spoofing to achieve state-of-the-art performance. Furthermore, models trained with auxiliary semantic information exhibit great generalizability~\cite{liu2020open} compared to other alternatives.

\setlength{\tabcolsep}{5pt}
\begin{table}[t]
\centering
\Large
\caption{The comparison of CelebA-Spoof with existing datasets of face anti-spoofing. Different illumination conditions and environments make up different sessions, (V means video, I means image; Ill. Illumination condition, Env. Environment; \textit{-} means this information is not annotated)}
\label{tab:Dataset}
\resizebox{\textwidth}{!}{
\begin{tabular}{@{}lcccccccc@{}}
\toprule
\multirow{2}{*}{Dataset} &
  \multirow{2}{*}{Year} &
  \multirow{2}{*}{Modality} &
  \multirow{2}{*}{\#Subjects} &
  \multirow{2}{*}{\#Data(V/I)} &
  \multirow{2}{*}{\#Sensor} &
  \multicolumn{3}{c}{\#Semantic Attribute} \\ \cmidrule{7-9} 
           &      &    &  &           &   & \#Face Attribute            & Spoof type & \#Session (Ill.,Env.) \\ \midrule
Replay-Attack~\cite{Replay-Attack} &
  2012 &
  RGB &
  50 &
  1,200 (V) &
  2 &\multirow{8}{*}{\textbackslash{}}  
   & 1 Print, 2 Replay & 1 (-.-) \\
CASIA-MFSD~\cite{CASIA-MFSD} & 2012 & RGB & 50   & 600 (V)   & 3 & & 1 Print, 1 Replay    & 3 (-.-)           \\
3DMAD~\cite{3DMAD} & 2014 & RGB/Depth  &14   & 255 (V)   & 2 & & 1 3D  mask    & 3 (-.-)           \\
MSU-MFSD~\cite{MSU-MFSD}   & 2015 & RGB & 35   & 440 (V)   & 2 &  & 1 Print, 2 Replay   & 1 (-.-)\\
Msspoof~\cite{msspoof-2015} & 2015 & RGB/IR & 21   & 4,704 (I)   & 2 &  & 1 Print   & 7 (-.7)\\
HKBU-MARs V2~\cite{HKBU-MARsV2}   & 2016 & RGB &12   & 1,008 (V)   & 7 &  & 2 3D masks   & 6 (6.-)   \\
MSU-USSA~\cite{MSU-USSA}   & 2016 & RGB & 1,140 & 10,260 (I) & 2 &  & 2 Print, 6 Replay & 1 (-.-)             \\
Oulu-NPU~\cite{oulu-NPU}   & 2017 &  RGB & 55   & 5,940 (V)  & 6 &  & 2 Print, 2 Replay   & 3 (-.-)          \\
SiW~\cite{SiW}        & 2018 & RGB & 165  & 4,620 (V)  & 2 & & 2 Print, 4 Replay & 4 (-.-)            \\
CASIA-SURF~\cite{casiasurf} & 2018 & RGB/IR/Depth & 1,000 & 21,000 (V) & 1  &   & 5 Paper Cut  & 1 (-.-)          \\
CSMAD~\cite{CSMAD} & 2018 & RGB/IR/Depth/LWIR &14 & 246 (V),17 (I) & 1  &   & 1 silicone mask  & 4 (4.-)          \\
HKBU-MARs V1+~\cite{Liu_2018_ECCV} & 2018 & RGB & 12 & 180(v) & 1  &   & 1 3D mask  & 1 (1.-)          \\
SiW-M~\cite{SiW-M} &
  2019 & 
  RGB &
  493 &
  1,628 (V) &
  4 &  &  \begin{tabular}[c]{@{}c@{}}1 Print, 1 Replay \\ 5 3D Mask, 3 Make Up, 3 Partial \end{tabular} & 3 (-.-) \\ \midrule
\textbf{CelebA-Spoof}&
  \textbf{2020} &
  \textbf{RGB} &
  \textbf{10,177} &
  \textbf{625,537 (I)} &
  \textbf{\textgreater{}10}  &
  \textbf{40}  &
 \textbf{\begin{tabular}[c]{@{}c@{}}3 Print, 3 Replay\\ 1 3D, 3 Paper Cut \end{tabular}} &
  \textbf{8 (4,2)} \\ \bottomrule[2pt]
\end{tabular}%
}
\end{table}

In summary, the \textbf{contributions} of this work are three-fold: \textbf{1)} We contribute a large-scale face anti-spoofing dataset, \textbf{CelebA-Spoof}, with 625,537 images from 10,177 subjects, which includes 43 rich attributes on face, illumination, environment and spoof types. \textbf{2)} Based on these rich attributes, we further propose a simple yet powerful multi-task framework, namely \textbf{AENet}. Through AENet, we conduct extensive experiments to explore the roles of semantic information and geometric information in face anti-spoofing. \textbf{3)} To support comprehensive evaluation and diagnosis, we establish three versatile benchmarks to evaluate the performance and generalization ability of various methods under different carefully-designed protocols. With several valuable observations revealed, we demonstrate the effectiveness of CelebA-Spoof and its rich attributes which can significantly facilitate future research.

\section{Related Work}

\noindent\textbf{Face Anti-Spoofing Datasets.}
Face anti-spoofing community mainly has three types of datasets. First, the multi-modal dataset: 3DMAD \cite{3DMAD}, Msspoof \cite{Msspoof}, CASIA-SURF \cite{casiasurf} and CSMAD~\cite{CSMAD}. However, since widespread used mobile phones are not equipped with suitable modules, such datasets cannot be widely used in the real scene. Second is the single-modal dataset, such as Replay Attack \cite{Replay-Attack}, CASIA-MFSD \cite{CASIA-MFSD}, MSU-MFSD \cite{MSU-MFSD}, MSU-USSA \cite{MSU-USSA} and HKBU-MARS V2~\cite{HKBU-MARsV2}. But these datasets have been collected for more than three years. With the rapid development of electronic equipment, the acquisition equipment of these datasets is completely outdated and cannot meet the actual needs. SiW \cite{SiW}, Oulu-NPU \cite{oulu-NPU} 
and HKBU-MAR V1+~\cite{Liu_2018_ECCV} are relatively up-to-date. However, the limited number of subjects, spoof types, and environment (only indoors) in these datasets does not guarantee for the generalization capability required in the real application. Third, SiW-M \cite{SiW-M} is mainly used for Zero-Shot face anti-spoofing tasks. CelebA-Spoof datasets have 625537 pictures from 10177 subjects, 8 scenes (2 environments * 4 illumination conditions) with rich annotations. The characteristic of Large-scale and diversity can further fill the gap between face anti-spoofing dataset and real scenes. with rich annotations we can better analyze face anti-spoofing task. All datasets mentioned above are listed in Table \ref{tab:Dataset}.

\noindent\textbf{Face Anti-Spoofing Methods.}
In recent years, face anti-spoofing algorithms have seen great progress. Most traditional algorithms focus on handcrafted features, such as LBP \cite{Replay-Attack,LBP_HoG:maatta2012face,LBP:ojala2002multiresolution,LBP_HoG:yang2013face}, HoG \cite{LBP_HoG:maatta2012face,LBP_HoG:yang2013face,HoG:schwartz2011face} and  SURF \cite{Surf_YCBCr:boulkenafet2016face}. Other works also focused on temporal features such as eye-blinking \cite{eyeblink:pan2007eyeblink,eyeblink:sun2007blinking} and lips motion \cite{lipmotion:kollreider2007real}. In order to improve the robustness to light changes, some researchers have paid attention to different color spaces, such as HSV \cite{HSV:boulkenafet2016face}, YCbcR \cite{Surf_YCBCr:boulkenafet2016face} and Fourier spectrum \cite{Fourier:li2004live}.
With the development of the deep learning model, researchers have also begun to focus on Convolutional Neural Network based methods. \cite{CNN_based:feng2016integration,CNN_based:li2016original} considered the face PAD problem as binary classification and perform good performance. The method of auxiliary supervision is also used to improve the performance of binary classification supervision. Atoum \textit{et al.} let the full convolutional network to learn the depth map and then assist the binary classification task. Liu \textit{et al.}~\cite{Liu_2018_ECCV,liu20163d} proposed  remote toplethysmography (rPPG signal)-based methods to foster the development of 3D face anti-spoofing. Liu \textit{et al.} \cite{SiW} proposed to leverage depth map combined with rPPG signal as the auxiliary supervision information. Kim \textit{et al.} \cite{kim2019basn} proposed using depth map and reflection map as the Bipartite auxiliary supervision. Besides, Yang \textit{et al.} \cite{STASN} proposed to combine the spatial information with the temporal information in the video stream to improve the generalization of the model. Amin \textit{et al.} \cite{jourabloo2018face} tackled the problem of face anti-spoofing by decomposing a spoof photo into a Live photo and a Spoof noise pattern. These methods mentioned above are prone to over-fitting on the training data, the generalization performance is poor in real scenarios. In order to solve the poor generalization problem, Shao \textit{et al.} \cite{deep_domain_generalization_2019_CVPR} adopted transfer learning to further improve performance. Therefore, a more complex face anti-spoofing dataset with large-scale and diversity is necessary. From extensive experiments, CelebA-Spoof has been shown to significantly improve generalization of basic models, In addition, based on auxiliary semantic information method can further achieve better generalization.

\section{CelebA-Spoof Dataset}
Existing face anti-spoofing datasets cannot satisfy the requirements for real scenario applications. As shown in Table~\ref{tab:Dataset}, most of them contain fewer than $200$ subjects and $5$ sessions, meanwhile they are only captured indoor with fewer than $10$ types of input sensors. 
On the contrary, our proposed CelebA-Spoof dataset provides $625,537$ pictures and $10,177$ subjects, therefore offering a superior comprehensive dataset for the area of face anti-spoofing. Furthermore, each image is annotated with $43$ attributes. This abundant information enrich the diversity and make face anti-spoofing more illustrative. To our best knowledge, our dataset surpasses all the existing datasets both in scale and diversity.

In this section, we describe our CelebA-Spoof dataset and analyze it through a variety of informative statistics. The dataset is built based on CelebA~\cite{CelebA}, where all the live people in this dataset are from CelebA. We collect and annotate Spoof images of CelebA-Spoof.

\subsection{Dataset Construction}
\label{subsec: dataset_construction}

\begin{figure}[t]
\centering
\includegraphics[width=0.95\textwidth]{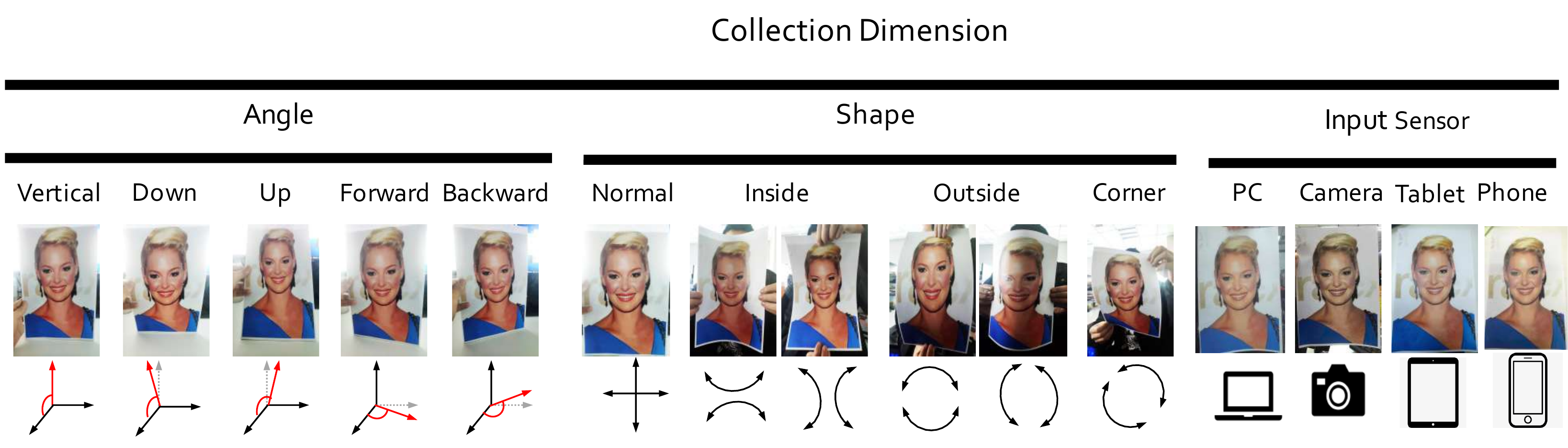}
\caption{An illustration of the collection dimension in CelebA-Spoof. In detail, these three dimensions boost the diversity of the dataset}
\label{fig:collection dimension}
\end{figure}

\noindent \textbf{Live Data.}
The \textit{live} data are directly inherited from CelebA dataset~\cite{CelebA}. CelebA is a well-known large-scale facial attribute dataset with more than eight million attribute labels. It covers face images with large pose variations and background clutters. We manually examine the images in CelebA and remove those ``spoof'' images, including posters, advertisements and cartoon portrait\footnote{There are $347$ images of this kind. Examples are shown in the supplementary material.}.

% spoof images
\noindent \textbf{Spoof Instrument Selection.}
The source for \textit{spoof} instruments is selected from the aforementioned live data. 
There are totally more than $202,599$ \textit{live} images from $10,177$ subjects. Each subject has multiple images ranging from 5 to 40. All subjects are covered in our spoof instrument production. In addition, to guarantee both the diversity and balance of spoof instruments, some subjects are filtered.
% 
% To guarantee the quality of spoof data, we choose images with face area larger than a threshold, since when we build ``facial mask'', the faces with too small sizes do not meet the common requirement. 
% 
Specifically, for one subject with more than $k$ source images, we rank them according to the face size with the bounding box provided by CelebA and select Top-$k$ source images.
For those subjects with fewer than $k$ source images, we directly adopt all of them. 
We set $k=20$. As a result, $87,926$ source images are selected from $202,599$ for further spoof instruments manufacture.

% collect spoof data
\noindent \textbf{Spoof Data Collection.}
We hired $8$ collectors to collect spoof data and another $2$ annotators to refine labeling for all data. To improve the generalization and diversity of the dataset, as shown in Figure~\ref{fig:collection dimension}, we define three collection dimensions with fine-grained quantities: 
1) \textit{Five Angles} - All spoof type need to traverse all five types of angles including ``vertical'', ``down'', ``up'', ``forward'' and ``backward''. The angle of inclination is between [$-30^{\circ}$, $30^{\circ}$].
2) \textit{Four Shapes} - There are a total of four shapes, \textit{i.e.}~``normal'', ``inside'', ``outside'' and ``corner''. 
3) \textit{Four Sensors} - We collected $24$ popular devices with four types,  \textit{i.e.}~``PC'', ``camera'', ``tablet'' and ``phone'', as the input sensors\footnote{Detailed information can be found in the supplementary material.}. These devices are equipped with different resolutions, ranging from 40 million to 12 million pixels. The number of input sensors is far more than the existing face anti-spoofing datasets as shown in Table \ref{tab:Dataset}.

\begin{figure}[t]
\centering
\includegraphics[width=0.95\textwidth]{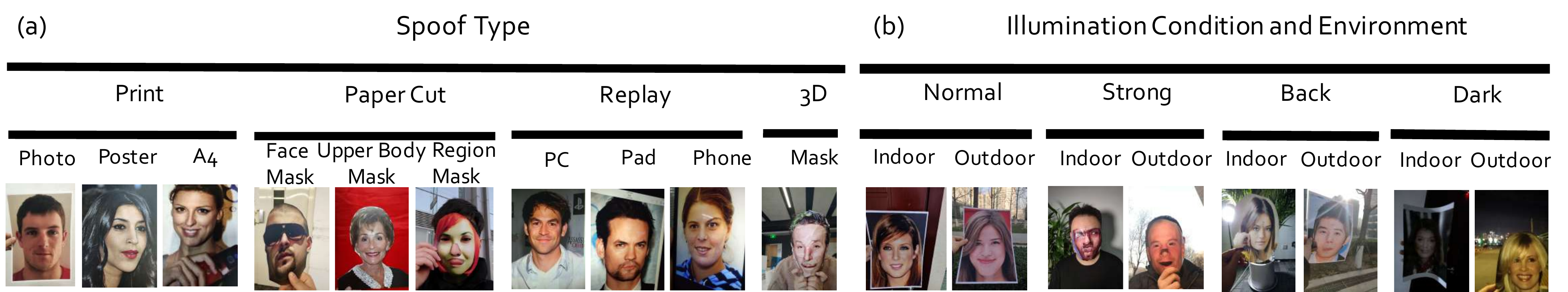}
\caption{Representative examples of the semantic attributes (\textit{i.e.}~spoof type, illumination and environment) defined upon spoof images. In detail, (a) 4 macro-types and 11 micro-types of spoof type and (b) 4 illumination and 2 types of environmental conditions are defined
}
\label{fig:attribute sample}
\end{figure}

\subsection{Semantic Information Collection}
\label{Attributes}
% 具体详细些各个attribute
% 
In recent decades, studies in attribute-based representations of objects, faces, and scenes have drawn large attention as a complement to categorical representations. However, rare works attempt to exploit semantic information in face anti-spoofing.
Indeed, for face anti-spoofing, additional semantic information can characterize the target images by attributes rather than discriminated assignment into a single category, \textit{i.e.}~``live'' or ``spoof''.

% ===== 1. ``live'' face attributes contribute to ``spoof'' images
\noindent\textbf{Semantic for Live - Face Attribute $\mathcal{S}^\text{f}$.} 
%\noident\textbf{Semantic for Live - Face Attribute $\mathcal{S}^\text{f}$.}  
In our dataset, we directly adopt $40$ types of face attributes defined in CelebA~\cite{CelebA} as ``live'' attributes. 
Attributes of ``live'' faces always refer to gender, hair color, expression and \textit{etc.} These abundant semantic cues have shown their potential in providing more information for face identification. It is the first time to incorporate them into face anti-spoofing. Extensive studies can be found in Sec.~\ref{sec:Study of semantic information}.

% ===== 2. ``spoof'' images own its specific attributes

\noindent\textbf{Semantic for Spoof - Spoof Type $\mathcal{S}^\text{s}$, Illumination $\mathcal{S}^\text{i}$, and Environment $\mathcal{S}^\text{e}$.} 
Differs to ``live'' face attributes, ``spoof'' images might be characterized by another bunch of properties or attributes as they are not only related to the face region. Indeed, the material of spoof type, illumination condition and environment where spoof images are captured can express more semantic information in ``spoof'' images, as shown in Figure~\ref{fig:attribute sample}.
Note that the combination of illumination and environment forms the ``session'' defined in the existing face anti-spoofing dataset. As shown in Table~\ref{tab:Dataset}, the combination of four illumination conditions and two environments forms 8 sessions. To our best knowledge, CelebA-Spoof is the first dataset covering spoof images in outdoor environment.

\begin{figure}[t]
\centering
\includegraphics[width=0.95\textwidth]{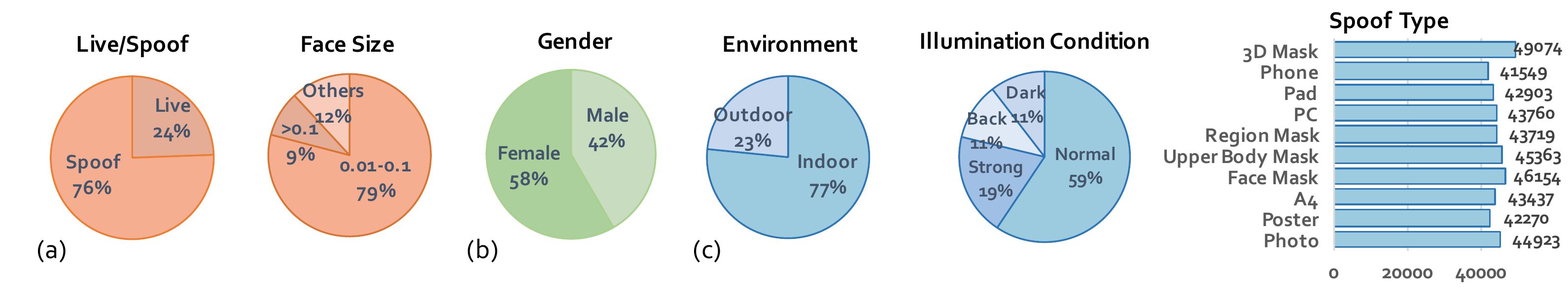}
\caption{The statistical distribution of CelebA-Spoof dataset. (a) Overall live and spoof distribution as well as the face size statistic. (b) An exemplar of live attribute, \textit{i.e.} ``gender''. (c) Three types of spoof attributes
}
\label{fig:Attribute statistics}
\end{figure}

\subsection{Statistics on CelebA-Spoof Dataset}

The CelebA-Spoof dataset is constructed with a total of $625,537$ images. As shown in~\ref{fig:Attribute statistics}(a), the ratio of live and spoof is 1 : 3. 
Face size in all images is mainly between 0.01 million pixels to 0.1 million pixels.  
We split the CelebA-Spoof dataset into training, validation, and test sets with a ratio of 8 : 1 : 1. Note that all three sets are guaranteed to have no overlap on subjects, which means there is no case of a live image of one certain subject in the training set while its counterpart spoof image in the test set.
The distribution of live images in three splits is the same as that defined in the CelebA dataset.

The semantic attribute statistics are shown in Figure~\ref{fig:Attribute statistics}(c). The portion of each type of attack is almost the same to guarantee a balanced distribution. It is easy to collect data under normal illumination in an indoor environment where most existing datasets adopt. Besides such easy cases, in CelebA-Spoof dataset, we also involve $12\%$ dark, $11\%$ back, and $19\%$ strong illumination. Furthermore, both indoor and outdoor environments contain all illumination conditions.

\section{Auxiliary Information Embedding Network}
\label{Auxiliary supervision based model}

\begin{figure}[t]
\centering
\includegraphics[width=0.8\textwidth]{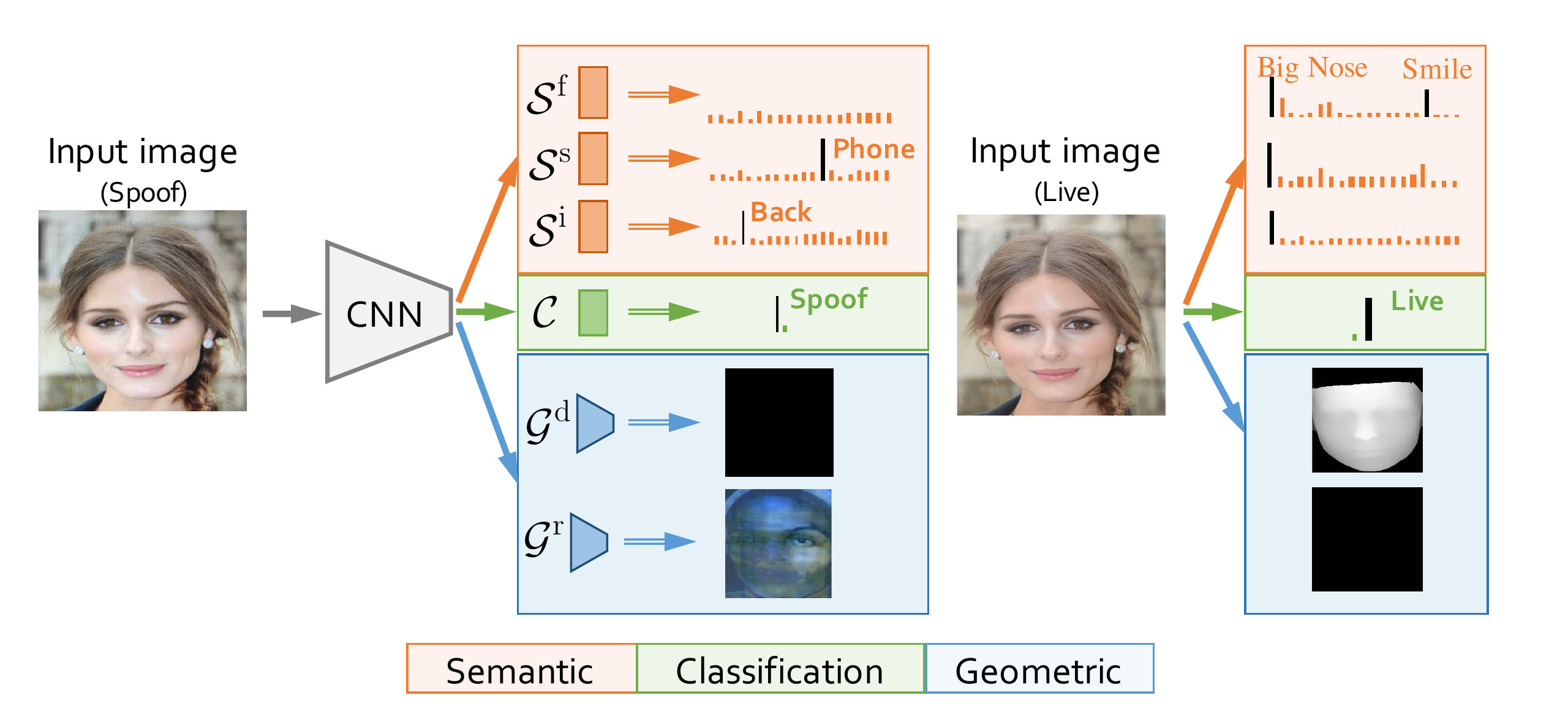}
\caption{\footnotesize{Auxiliary information Embedding Network (AENet). We use two \texttt{Conv$_{3\times 3}$} after CNN and upsample to size $14\times 14$ to learn the geometric information. Besides, we use three  \texttt{FC} layers to learn the semantic information. The prediction score of $\mathcal{S}^\text{f}$ of spoof image should be very low and the prediction result of $\mathcal{S}^\text{s}$ and $\mathcal{S}^\text{i}$ of live image should be ``No illumination'' and ``No attack'' which belongs to the first label in $\mathcal{S}^\text{s}$ and $\mathcal{S}^\text{i}$}}
\label{fig:CNN}
\end{figure}

Equipped with CelebA-Spoof dataset, in this section, we design a simple yet effective network named \textbf{A}uxiliary information \textbf{E}mbedding \textbf{N}etwork (AENet), as shown in Figure~\ref{fig:CNN}. 
In addition to the main binary classification branch (in green), we \textbf{1)} Incorporate the \textit{semantic} branch (in orange) to exploit the auxiliary capacity of rich annotated semantic attributes in the dataset, and \textbf{2)} Benchmark the existing \textit{geometric} auxiliary information within this unified multi-task framework. 

\noindent\textbf{AENet$_{\mathcal{C},\mathcal{S}}$.} 
refers to the multi-task jointly learn auxiliary ``semantic'' attributes and binary ``classification'' labels. Such auxiliary semantic attributes defined in our dataset provide complement cues rather than discriminated assignment into a single category.
The semantic attributes are learned via the backbone network followed by three \texttt{FC} layers. In detail, given a batch of $n$ images, based on AENet$_{\mathcal{C},\mathcal{S}}$, we learn live/spoof class $\{\mathcal{C}_{k}\}_{k=1}^{n}$ and semantic information, \textit{i.e.} live face attributes $\{\mathcal{S}_{k}^{\text{f}}\}_{k=1}^{n}$, spoof type $\{\mathcal{S}_{k}^{\text{s}}\}_{k=1}^{n}$ and illumination conditions $\{\mathcal{S}_{k}^{\text{i}}\}_{k=1}^{n}$ simultaneously\footnote{Note that we do not learn environments $\mathcal{S}^{\text{e}}$ since we take face image as input where environment cues (\textit{i.e.}~indoor or outdoor) cannot provide more valuable information yet illumination influences much.}. The loss function of our AENet$_{\mathcal{C},\mathcal{S}}$ is 
\begin{equation}
\mathcal{L}_{c,s}=\mathcal{L}_{\mathcal{C}} + \lambda_{f}\mathcal{L}_{\mathcal{S}^\text{f}} + \lambda_{s}\mathcal{L}_{\mathcal{S}^\text{s}} + \lambda_{i}\mathcal{L}_{\mathcal{S}^\text{i}},
\end{equation}
where $\mathcal{L}_{\mathcal{S}^\text{f}}$ is binary cross entropy loss. $\mathcal{L}_{\mathcal{C}}$, $\mathcal{L}_{\mathcal{S}^\text{s}}$ and $\mathcal{L}_{\mathcal{S}^\text{i}}$ are softmax cross entropy losses. We set the loss weights $\lambda_{f} = 1$, $\lambda_{s} = 0.1$ and $\lambda_{i} = 0.01$, $\lambda$ values are empirically selected to balance the contribution of each loss.

\noindent\textbf{AENet$_{\mathcal{C},\mathcal{G}}$.} 
Besides the semantic auxiliary information, some recent works claim some geometric cues such as \textit{reflection map} and \textit{depth map} can facilitate face anti-spoofing. As shown in Figure~\ref{fig:CNN} (marked in blue), spoof images exhibit even and the flat surfaces which can be easily distinguished by the depth map. The reflection maps, on the other hand, may display reflection artifacts caused by reflected light from flat surface. However, rare works explore their pros and cons.

AENet$_{\mathcal{C},\mathcal{G}}$ also learn auxiliary geometric information in a multi-task fashion with live/spoof classification. Specifically, we concate a \texttt{Conv}$\_{3\times 3}$ after the backbone network and upsample to $14\times 14$ to output the geometric maps. We denote depth and reflection cues as $\mathcal{G}^\text{d}$ and $\mathcal{G}^\text{r}$ respectively. The loss function is defined as
\begin{equation}
\mathcal{L}_{c,g}=\mathcal{L}_{c} + \lambda _{d}\mathcal{L}_{\mathcal{G}^\text{d}} + \lambda_{r}\mathcal{L}_{\mathcal{G}^\text{r}},
\end{equation}
where $\mathcal{L}_{\mathcal{G}^\text{d}}$ and $\mathcal{L}_{\mathcal{G}^\text{r}}$ are mean squared error losses. $\lambda_{d}$ and $\lambda_{r}$ are set to $0.1$. In detail, refer to~\cite{kim2019basn}, the ground truth of the depth map of live image is generated by PRNet~\cite{PRNet} and the ground truth of the reflection map of the spoof image is generated by the method in~\cite{refletion_map}. Besides, the ground truth of the depth map of the spoof image and the ground truth of the reflection map of the live images are zero.

\section{Experimental Settings}

\noindent\textbf{Evaluation Metrics.}
Different metrics have been taken to evaluate previous methods that make the comparison inconsistent. To establish a comprehensive benchmark, we unify all the commonly used metrics (\textit{i.e.}~APCER, BPCER, ACER, EER, and HTER)\footnote{Detailed definitions and formulations are listed in the supplementary material.}~\cite{oulu-NPU,SiW,jourabloo2018face,STASN} and add another two criteria (\textit{i.e.}~FPR@Recall and AUC). 
APCER and BPCER are used to evaluate the error rate of live and spoof image respectively. ACER is the average of the APCER and the BPCER.
Besides, AUC can evaluate the overall classification performance, and FPR@Recall can expose detailed Recalls corresponding to some specific FPRs.
The aforementioned metrics are employed on intra-dataset (CelebA-Spoof) evaluation, and for cross-dataset evaluation, HTER~\cite{LBP_TOP} is used extensively.

\noindent\textbf{Implementation Details.}
We initialize the backbone network\footnote{For fair comparison, ResNet-18 is used in all experiments. We also take another heavier backbone, \textit{i.e.}~Xception, to enrich the benchmarks.} with the parameters pre-trained on ImageNet. 
The network takes face image as the input with a size of $224\times 224$. The bounding box of faces are extracted by  RetinaFace~\cite{deng2019retinaface}. We use color distortion for data augmentation.
SGD optimizer is adopted for training. The learning rate is set to 0.005 for 50 epochs.

\section{Ablation Study on CelebA-Spoof}
\label{sec:experiments}
Based on our rich annotations in CelebA-Spoof and the designed AENet, we conduct extensive experiments to analyze semantic information and geometric information. Several valuable observations have been revealed: \textbf{1)} We validate that $\mathcal{S}^\text{f}$ and $\mathcal{S}^\text{s}$ can facilitate live/spoof classification performance greatly. \textbf{2)} We analyze the effectiveness of geometric information on different spoof types and find that depth information is particularly sensitive to dark illumination.

\setlength{\tabcolsep}{5pt}
\begin{table}[t]
\centering
\ra{1.3}
\LARGE
\caption{Different settings in ablation study. For Baseline, we use softmax score of $\mathcal{C}$ for classification (a) For AENet$_\mathcal{S}$, we use the average softmax score of $\mathcal{S}^\text{f}$, $\mathcal{S}^\text{s}$ and $\mathcal{S}^\text{i}$ for classification. AENet$_{\mathcal{S}^\text{f}}$, AENet$_{\mathcal{S}^\text{s}}$ and AENet$_{\mathcal{S}^\text{i}}$ refer to each single spoof semantic attribute respectively. Based on AENet$_{\mathcal{C},\mathcal{S}}$, w/o $\mathcal{S}^\text{f}$,  w/o $\mathcal{S}^\text{s}$,  w/o $\mathcal{S}^\text{i}$ mean AENet$_{\mathcal{C},\mathcal{S}}$ discards $\mathcal{S}^\text{f}$, $\mathcal{S}^\text{s}$ and $\mathcal{S}^\text{i}$ respectively. (b) For AENet$_{\mathcal{G}^\text{d}}$, we use $\left \|\mathcal{G}^\text{d}\right \|_{2}$ for classification. Based on AENet$_{\mathcal{C},\mathcal{G}}$, w/o $\mathcal{G}^\text{d}$,  w/o $\mathcal{G}^\text{r}$ mean AENet$_{\mathcal{C},\mathcal{G}}$ discards $\mathcal{G}^\text{d}$ and $\mathcal{G}^\text{r}$ respectively}
\label{tab:table for semantic information and geometric information}
\resizebox{\textwidth}{!}{%
\begin{tabular}{@{}c|c|cccc|cccc||c|c|c|ccc@{}}
\hline
                            (a) & Baseline   & AENet$_\mathcal{S}$ & AENet$_{\mathcal{S}^\text{f}}$ &
                         AENet$_{\mathcal{S}^\text{s}}$ &
                         AENet$_{\mathcal{S}^\text{i}}$ & \begin{tabular}[c]{@{}c@{}}AENet$_{\mathcal{C},\mathcal{S}}$\\ w/o $\mathcal{S}^\text{f}$ \end{tabular} & \begin{tabular}[c]{@{}c@{}}AENet$_{\mathcal{C},\mathcal{S}}$\\ w/o $\mathcal{S}^\text{s}$ \end{tabular} &  \begin{tabular}[c]{@{}c@{}}AENet$_{\mathcal{C},\mathcal{S}}$\\ w/o $\mathcal{S}^\text{i}$ \end{tabular} & AENet$_{\mathcal{C},\mathcal{S}}$   & (b) & Baseline & AENet$_{\mathcal{G}^\text{d}}$ &
 \begin{tabular}[c]{@{}c@{}}AENet$_{\mathcal{C},\mathcal{G}}$\\ w/o $\mathcal{G}^\text{r}$ \end{tabular} & \begin{tabular}[c]{@{}c@{}}AENet$_{\mathcal{C},\mathcal{S}}$\\ w/o $\mathcal{G}^\text{d}$ \end{tabular} &  AENet$_{\mathcal{C},\mathcal{G}}$ \\ \hline
Live/Spoof    &  $\surd$               &    &  &  &    & $\surd$     & $\surd$     & $\surd$ & $\surd$ & Live/Spoof &   $\surd$&   &  $\surd$    & $\surd$ & $\surd$  \\
Face Attribute    &            & $\surd$    &  $\surd$ &  &     &       & $\surd$     & $\surd$ & $\surd$ & \multirow{2}{*}{Reflection Map} &  &  &   &  \multirow{2}{*}{$\surd$}  &  \multirow{2}{*}{$\surd$}  \\
Spoof Type & & $\surd$  & & $\surd$ &   & $\surd$    &       & $\surd$ & $\surd$ & & &  &  &   \\
Illumination Conditions   &     & $\surd$ &         &         & $\surd$ & $\surd$ & $\surd$ &  & $\surd$ & Depth map &  & $\surd$     & $\surd$ &  & $\surd$  \\ 
\hline
\end{tabular}%
}
\end{table}

\subsection{Study of Semantic Information}
\label{sec:Study of semantic information}
In this subsection, we explore the role of different semantic informations annotated in CelebA-Spoof on face anti-spoofing. Based on AENet$_{\mathcal{C},\mathcal{S}}$, we design eight different models in the Table \ref{tab:table for semantic information and geometric information}(a). The \textit{key} observations are:

\noindent\textbf{Binary Supervision is Indispensable.}
As shown in Table \ref{tab:multi_label_resnet}(a), Compared to baseline, AENet$_\mathcal{S}$ which only leverages three semantic attributes to do the auxiliary job cannot surpass the performance of baseline. However, as shown in \ref{tab:multi_label_resnet}(b), AENet$_{\mathcal{C},\mathcal{S}}$ which jointly learns auxiliary semantic attributes and binary classification significantly improves the performance of baseline. Therefore we can infer that even such rich semantic information cannot fully replace live/spoof information. But live/spoof with semantic attributes as auxiliary information can be more effective. This is because the semantic attributes of an image cannot be included completely, and a better classification performance cannot be achieved only by relying on several annotated semantic attributes. However, semantic attributes can help the model pay more attention to cues in the image, thus improving the classification performance of the model.

\begin{figure}[t]
\centering
\includegraphics[width=0.9\textwidth]{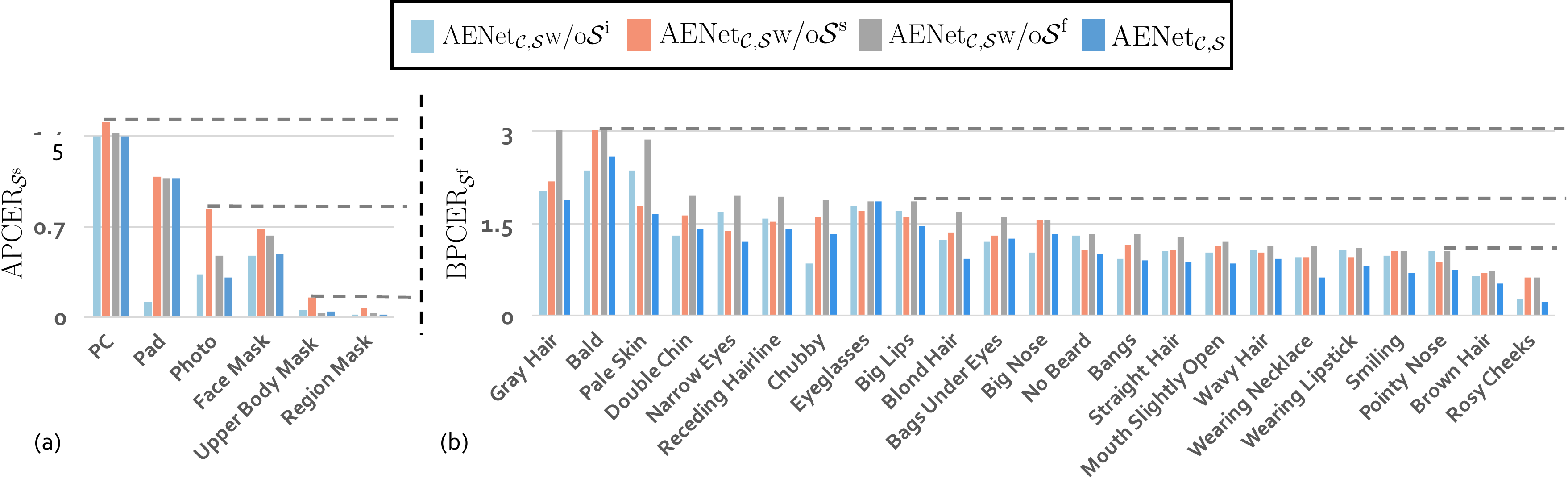}
\caption{Representative examples of dropping partial semantic attributes on AENet$_{\mathcal{C},\mathcal{S}}$ performance. In detail, higher APCER$_{\mathcal{S}^\text{s}}$ and BPCER$_{\mathcal{S}^\text{f}}$ are worse results. (a) Spoof types where AENet$_{\mathcal{C},\mathcal{S}}$ w/o $\mathcal{S}^\text{s}$ achieve the worst APCER$_{\mathcal{S}^\text{s}}$. (b) Face attributes where AENet$_{\mathcal{C},\mathcal{S}}$ w/o $\mathcal{S}^\text{f}$ achieve the worst BPCER$_{\mathcal{S}^\text{f}}$}
\label{fig:Attack APCER}
\end{figure}

\noindent\textbf{Semantic Attribute Matters.}
From Table \ref{tab:multi_label_resnet}(c), we study the impact of different individual semantic attributes on AENet$_{\mathcal{C},\mathcal{S}}$. As shown in this table, AENet$_{\mathcal{C},\mathcal{S}}$ w/o $\mathcal{S}^\text{s}$ achieves the worst APCER. Since APCER reflects the classification ability of spoof images, it shows that compared to other semantic attributes, \textit{spoof types} would significantly affect the performance of the spoof images classification of AENet$_{\mathcal{C},\mathcal{S}}$. 
Furthermore, we list detail information of AENet$_{\mathcal{C},\mathcal{S}}$
% \footnote{Calculating APCER over each spoof type.}
in Figure \ref{fig:Attack APCER}(a). As shown in this figure,  AENet$_{\mathcal{C},\mathcal{S}}$ without \textit{spoof types} gets the 5 worst APCER$_{\mathcal{S}^\text{s}}$ out of 10 APCER$_{\mathcal{S}^\text{s}}$ and we show up these 5 values in this figure. 
Besides, in Table \ref{tab:multi_label_resnet}(b), AENet$_{\mathcal{C},\mathcal{S}}$ w/o $\mathcal{S}^\text{f}$ gets the highest BPCER. And we also obtain the BPCER$_{\mathcal{S}^\text{f}}$ of each face attribute. As shown in Figure \ref{fig:Attack APCER}(b), among 40 face attributes, BPCER$_{\mathcal{S}^\text{f}}$ of AENet$_{\mathcal{C},\mathcal{S}}$ w/o $\mathcal{S}^\text{f}$ occupies 25 worst scores. Since BPCER reflects the classification ability of live images, it demonstrate $\mathcal{S}^\text{f}$ plays an important role in the classification of live images.

\setlength{\tabcolsep}{5pt}
\begin{table}[t]
\centering
\caption{Semantic information study results in Sec. \ref{sec:Study of semantic information}. (a) AENet$_\mathcal{S}$ which only depends on semantic attributes for classification cannot surpass the performance of baseline. (b) AENet$_{\mathcal{C},\mathcal{S}}$ which leverages all semantic attributes achieve the best result. \textbf{Bolds} are the best results; $\uparrow$ means bigger value is better; $\downarrow$ means smaller value is better}
\label{tab:multi_label_resnet}
\resizebox{\textwidth}{!}{%
\begin{tabular}{@{}ccccccccccc@{}}
\toprule
\multirow{2}{*}{} &
\multirow{2}{*}{Model} &
  \multicolumn{3}{c}{Recall (\%)$\uparrow$} &
  \multirow{2}{*}{AUC $\uparrow$} &
  \multirow{2}{*}{EER (\%) $\downarrow$} &
  \multirow{2}{*}{APCER (\%) $\downarrow$} &
  \multirow{2}{*}{BPCER (\%) $\downarrow$} &
  \multirow{2}{*}{ACER (\%) $\downarrow$} \\ \cmidrule{3-5}
    &    &FPR = 1\% & FPR = 0.5\%  & FPR = 0.1\%\     &    &    &                 &                &                                             \\ \midrule
\multirow{2}{*}{(a)} & Baseline& 97.9          & 95.3          & 85.9          & 0.9984          & 1.6          & 6.1           & 1.6           & 3.8           \\ %\hline
 & AENet$_\mathcal{S}$          & 98.0          & 96.0          & 80.4          & 0.9981          & 1.4          & 6.89          & 1.44          & 4.17          \\ \midrule
(b) & AENet$_{\mathcal{C},\mathcal{S}}$           & \textbf{98.8} & \textbf{97.4}          & \textbf{90.0}          & \textbf{0.9988} & \textbf{1.1} & \textbf{4.62}          & \textbf{1.09}          & \textbf{2.85}          \\\midrule
 \multirow{3}{*}{(c)} & AENet$_{\mathcal{C},\mathcal{S}}$ w/o $\mathcal{S}^\text{i}$            & 98.1          & 96.5          & 86.4          & 0.9982          & 1.3          & \textbf{4.62}          & 1.35          & 2.99          \\ %\hline
& AENet$_{\mathcal{C},\mathcal{S}}$ w/o $\mathcal{S}^\text{s}$           & 98.2          & 96.5          & 89.4          & 0.9986          & 1.3          & 5.31          & 1.25          & 3.28          \\ %\hline
& AENet$_{\mathcal{C},\mathcal{S}}$ w/o $\mathcal{S}^\text{f}$           & 97.8         & 95.4          & 83.6          & 0.9979          & 1.3        & 5.19          & 1.37          & 3.28          \\ %\hline
 \bottomrule% \hline
\end{tabular}%
}
\end{table}

\begin{figure}[t]
\centering
\includegraphics[width=\textwidth]{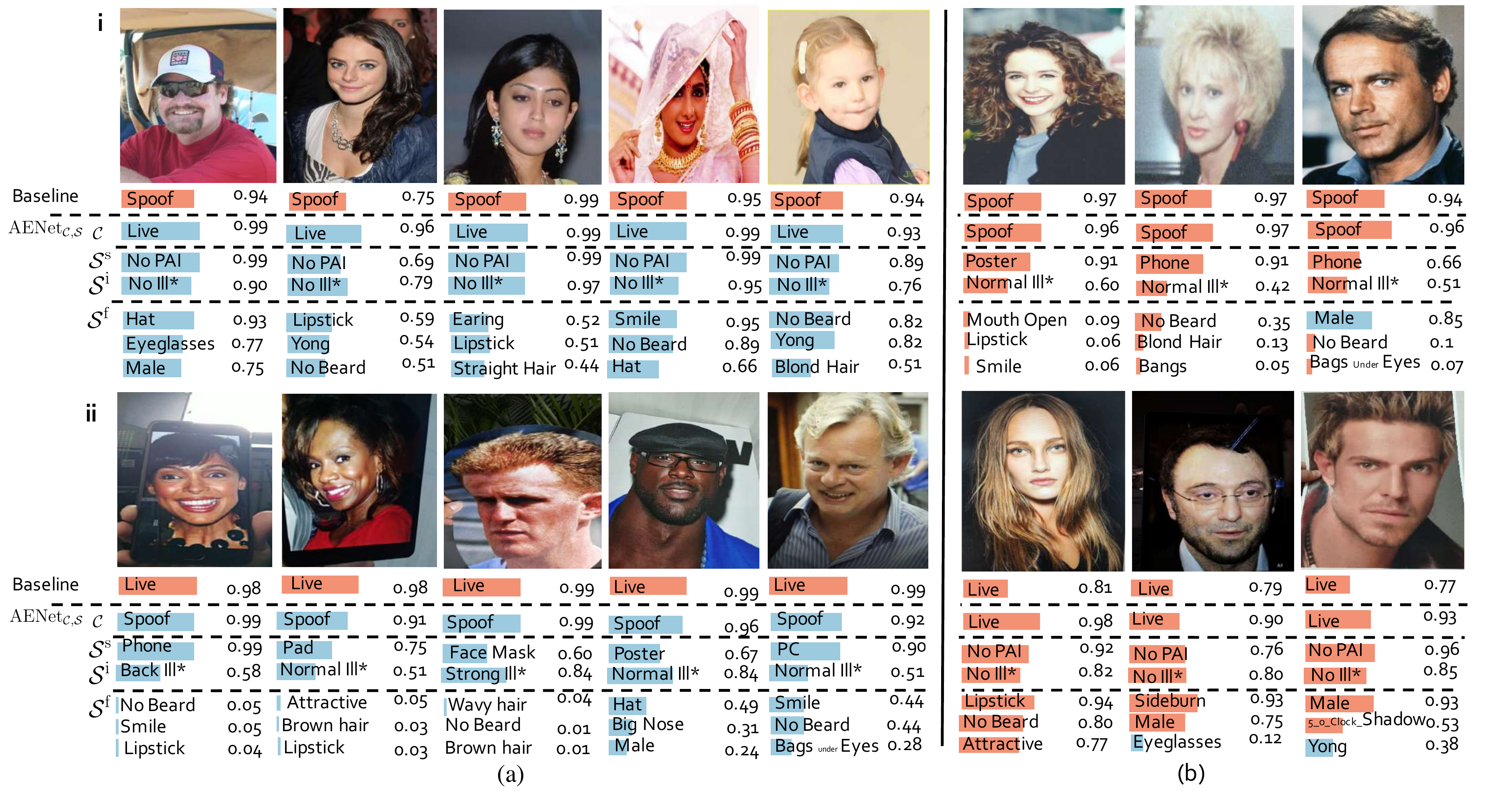}
\caption{Success and failure cases. The row(i) present the live image and row(ii) present the spoof image. For each image, the first row is the highest score of live/spoof prediction of baseline and others are the highest live/spoof and the highest semantic attributes predictions of AENet$_{\mathcal{C},\mathcal{S}}$. Blue indicates correctly predicted results and orange indicates the wrong results. In detail, we list the top three prediction scores of face attributes in the last three rows of each image}
\label{fig:fig2}
\end{figure}

\noindent\textbf{Qualitative Evaluation.}
Success and failure cases on live/spoof and semantic attributes predictions are shown in Figure \ref{fig:fig2}. For \textit{live} examples, the first example in Figure \ref{fig:fig2}(a-i) with ``glasses`` and ``hat`` help AENet$_{\mathcal{C},\mathcal{S}}$ to pay more attention to the clues of the live image and further improve the performance of prediction of live/spoof. Besides, the first example in Figure \ref{fig:fig2}(a-ii). AENet$_{\mathcal{C},\mathcal{S}}$ significantly improve the classification performance of live/spoof comparing to baseline. This is because spoof semantic attributes including ``back illumination'' and ``phone'' help AENet$_{\mathcal{C},\mathcal{S}}$ recognize the distinct characteristics of spoof image. Note that the prediction of the second example in Figure \ref{fig:fig2}(b-i) is mistaken.

\setlength{\tabcolsep}{5pt}
\begin{table}[t]
\centering
\caption{Geometric information study results in Sec. \ref{sec:Study of geometric information}. (a) AENet$_{\mathcal{G}^\text{d}}$ which only depends on the depth map for classification performs worst than baseline. (b) AENet$_{\mathcal{C},\mathcal{G}}$ which leverages all semantic attributes achieve the best result. \textbf{Bolds} are the best results; $\uparrow$ means bigger value is better; $\downarrow$ means smaller value is better}
\label{tab:Result of geometric}
\resizebox{\textwidth}{!}{%
\begin{tabular}{@{}cccccccccc@{}}
\toprule
\multirow{2}{*}{} &
\multirow{2}{*}{Model} &
  \multicolumn{3}{c}{Recall (\%)$\uparrow$} &
  \multirow{2}{*}{AUC$\uparrow$} &
  \multirow{2}{*}{EER (\%)$\downarrow$} &
  \multirow{2}{*}{APCER (\%)$\downarrow$} &
  \multirow{2}{*}{BPCER (\%)$\downarrow$} &
  \multirow{2}{*}{ACER (\%)$\downarrow$} \\ \cmidrule{3-5}
     &   &FPR = 1\% & FPR = 0.5\% & FPR = 0.1\% &               &      &      &  &    \\ \midrule
\multirow{2}{*}{(a)} &Baseline&  97.9      & 95.3        & 85.9        & 0.9984 & 1.6 & 6.1  & 1.6  & 3.8  \\ %\hline
& AENet$_{\mathcal{G}^\text{d}}$ & 97.8      & 96.2        & 87.0        & 0.9946 & 1.6 & 7.33 & 1.68 & 4.51 \\ \midrule
(b) & AENet$_{\mathcal{C},\mathcal{G}}$ & \textbf{98.4}      & \textbf{96.8}        & 86.7        & \textbf{0.9985} & \textbf{1.2} & \textbf{5.34} & \textbf{1.19} & \textbf{3.26} \\ \midrule
\multirow{2}{*}{(c)} & AENet$_{\mathcal{C},\mathcal{G}}$ w/o $\mathcal{G}^\text{d}$ & 98.3      & 96.1        & \textbf{87.7}        & 0.9976 & \textbf{1.2} & 5.91 & 1.27 & 3.59 \\ %\hline
 & AENet$_{\mathcal{C},\mathcal{G}}$ w/o $\mathcal{G}^\text{r}$ & 97.9      & 95.7        & 84.1        & 0.9973 & 1.3 & 5.71 & 1.38 & 3.55 \\  \bottomrule

\end{tabular}%
}
\end{table}

\subsection{Study of Geometric Information}
\label{sec:Study of geometric information}
\begin{figure}[t]
\centering
\includegraphics[width=0.95\textwidth]{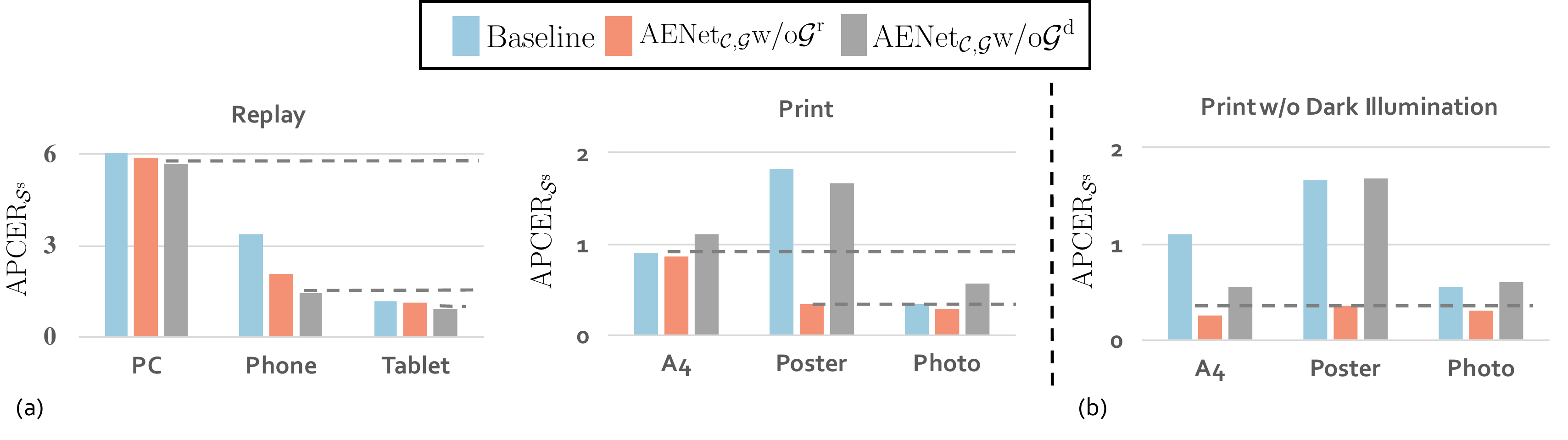}
\caption{Representative examples of the effectiveness of geometric information. Higher APCER$_{\mathcal{S}^\text{s}}$ is worse. (a) AENet$_{\mathcal{C},\mathcal{G}}$ w/o $\mathcal{G}^\text{d}$ perform the best in spoof type: ``replay''(macro definition) and AENet$_{\mathcal{C},\mathcal{G}}$ w/o $\mathcal{G}^\text{r}$ perform the best in spoof type: ``print''(macro definition). (b) The performance of AENet$_{\mathcal{C},\mathcal{G}}$ w/o $\mathcal{G}^\text{r}$ improve largely on spoof type: ``A4'', if we only calculate APCER under illumination conditions: ``normal'', ``strong'' and ``back''}
\label{fig:Depth APCER}
\end{figure}

Based on AENet$_{\mathcal{C},\mathcal{g}}$ under different settings, we design four models as shown in Table~\ref{tab:table for semantic information and geometric information}(b) and use semantic attributes we annotated to analyze the usage of geometric information in face anti-spoofing task. The \textit{key} observations are:

\noindent\textbf{Depth Maps are More Versatile.} As shown in Table \ref{tab:Result of geometric}(a), geometric information is insufficient to be the unique supervision for live/spoof classification. However, it can boost the performance of the baseline when it serves as an auxiliary supervision. Besides, we study the impact of different individual geometric information on AENet$_{\mathcal{C},\mathcal{G}}$ performance. As shown in Figure \ref{fig:Depth APCER}(a), AENet$_{\mathcal{C},\mathcal{G}}$ w/o $\mathcal{G}^\text{d}$ performs the best in spoof type: ``replay'' (macro definition), because the reflect artifacts appear frequently in these three spoof types. For ``phone'', AENet$_{\mathcal{C},\mathcal{G}}$ w/o $\mathcal{G}^\text{d}$ improves 56$\%$ comparing to the baseline. However AENet$_{\mathcal{C},\mathcal{G}}$ w/o $\mathcal{G}^\text{d}$ gets worse result than baseline in spoof type: ``print'' (macro definition). Moreover, AENet$_{\mathcal{C},\mathcal{G}}$ w/o $\mathcal{G}^\text{r}$ helps greatly to improve the classification performance of baseline in both ``replay'' and ``print''(macro definition). Especially for ``poster'', AENet$_{\mathcal{C},\mathcal{G}}$ w/o $\mathcal{G}^\text{r}$ improves baseline by 81$\%$. Therefore, the depth map can improve classification performance in most spoof types, but the function of the reflection map is mainly reflected in ``replay''(macro definition).

% Xception& 97.37     & 96.03       & 87.79       & \textbf{0.9986} & 0.017 & 5.17 & 1.78 & 3.48 \\ %\hline
% $Model_{G5}$ & 98.3      & \textbf{97.2}        & \textbf{91.4}        & 0.9982 & \textbf{0.012} & \textbf{4.98} & \textbf{1.26} & \textbf{3.12} \\

\noindent\textbf{Sensitive to Illumination.}
As shown in Figure \ref{fig:Depth APCER}(a), in spoof type ``print''(macro definition), the performance of the AENet$_{\mathcal{C},\mathcal{G}}$ w/o $\mathcal{G}^\text{r}$ on ``A4'' is much worse than ``poster'' and ``photo'', although they are both in ``print'' spoof type. The main reason for the large difference in performance among these three spoof types for AENet$_{\mathcal{C},\mathcal{G}}$ w/o $\mathcal{G}^\text{r}$ is that the learning of the depth map is sensitive to dark illumination, as shown in Figure \ref{fig:Depth APCER}(b). 
% Since the dark illumination share of poster is small,  AENet$_{\mathcal{C},\mathcal{G}}$w/o $\mathcal{G}^{r}$ perform well at it. However, the dark illumination share of A4 is large, AENet$_{\mathcal{C},\mathcal{G}}$w/o $\mathcal{G}^{r}$ perform poorly.  Moreover, 
When we calculate APCER under other illumination conditions: normal, strong and back, AENet$_{\mathcal{C},\mathcal{G}}$ w/o $\mathcal{G}^\text{r}$ achieves almost the same results among ``A4'', ``poster'' and ``photo''.

\section{Benchmarks}
\label{sec:benchmark}
% In addition to our large-scale and annotated dataset, we establish a series of competitive benchmarks for face anti-spoofing. 
In order to facilitate future research in the community, we carefully build three different benchmarks to investigate face anti-spoofing algorithms. Specifically, for a comprehensive evaluation, besides ResNet-18, we also provide the corresponding results based on a heavier backbone, \textit{i.e.}~Xception. Detailed information of the results based on Xception are shown in the supplementary material.

\subsection{Intra-Dataset Benchmark}

Based on this benchmark, models are trained and evaluated on the whole training set and testing set of CelebA-Spoof. This benchmark evaluates the overall capability of the classification models. According to different input data types, there are two kinds of face anti-spoof methods, \textit{i.e.} `` video-driven methods'' and ``image-driven methods''. Since the data in CelebA-Spoof are image-based, we benchmark state-of-the-art ``image-driven methods'' in this subsection.  
% two kinds of method benchmark existing state-of-the-art method and AENet$_{C,G}$, AENet$_{C,S}$ and AENet$_{C,G,S}$ (Combine geometric information and semantic information). Regardless of the amount of parameters, the state-of-the-art method on Oulu-NPU and SiW is \textit{BASN} \cite{kim2019basn}. This method designs its own network structure based on VGG16. The comparison results are shown in the Table \ref{tab:BASN}.Since semantic information can provide richer information than geometric information (42 semantic information and 2 geometric information), the AENet$_{C,S}$ achieved better results than AENet$_{C,G}$. Furthermore, 
%
As shown in Table \ref{tab:BASN}, AENet$_{\mathcal{C},\mathcal{S}, \mathcal{G}}$ which combines geometric and semantic information has achieved the best results on CelebA-Spoof. Specifically, our approach outperforms the state-of-the-art by 38\% with much fewer parameters.

\setlength{\tabcolsep}{5pt}
\begin{table}[t]
\centering
\LARGE
\caption{Intro-dataset Benchmark results on CelebA-Spoof. AENet$_{\mathcal{C},\mathcal{S}, \mathcal{G}}$ achieved the best result. \textbf{Bolds} are the best results; $\uparrow$ means bigger value is better; $\downarrow$ means smaller value is better. * Model 2 defined in \textit{Auxiliary} can be used as ``image driven method''}
\label{tab:BASN}
\resizebox{\textwidth}{!}{%
\begin{tabular}{@{}ccccccccccc@{}}
\toprule 
\multirow{2}{*}{Model} &
\multirow{2}{*}{Backbone} &
\multirow{2}{*}{Parm. (MB)} &
  \multicolumn{3}{c}{Recall (\%)$\uparrow$ } &
  \multirow{2}{*}{AUC$\uparrow$ } &
  \multirow{2}{*}{EER (\%)$\downarrow$ } &
  \multirow{2}{*}{APCER (\%)$\downarrow$ } &
  \multirow{2}{*}{BPCER (\%)$\downarrow$ } &
  \multirow{2}{*}{ACER (\%)$\downarrow$ } \\ \cmidrule{4-6}
         &   &   & FPR = 1\% & FPR = 0.5\%   & FPR = 0.1\%   &        &       &      &      &      \\ \midrule
Auxiliary*~\cite{SiW} & -  & 22.1 &
97.3      & 95.2          & 83.2          & 0.9972 & 1.2 & 5.71 & 1.41 & 3.56 \\ 
BASN \cite{kim2019basn} & VGG16 & 569.7 &
  98.9 &
  \textbf{97.8} &
  \textbf{90.9} &
  \textbf{0.9991} &
  1.1 &
  4.0 &
  1.1 &
  2.6 \\ %\hline
AENet$_{\mathcal{C},\mathcal{S}, \mathcal{G}}$ & ResNet-18   & 42.7        & 
\textbf{98.9} & 
97.3          & 
87.3          & 
0.9989 & 
\textbf{0.9} & 
\textbf{2.29}          & 
\textbf{0.96}         & 
\textbf{1.63}          \\ \bottomrule %\hline
\end{tabular}%
}
\end{table}

\subsection{Cross-Domain Benchmark}
 Since face anti-spoofing is an open-set problem, even though CelebA-Spoof is equipped with diverse images, it is impossible to cover all spoof types, environments, sensors, \textit{etc.} that exist in the real world. Inspired by~\cite{oulu-NPU,SiW}, we carefully design two protocols for CelebA-Spoof based on real-world scenarios. In each protocol, we evaluate the performance of trained models under controlled domain shifts. Specifically, we define two protocols. \textbf{1)} \textit{Protocol 1 - } Protocol 1 evaluates the cross-medium performance of various spoof types. This protocol includes 3 macro types of spoof, where each covers 3 micro types of spoof. These three macro types of spoof are ``print'', ``repay'' and ``paper cut''. In detail, in each macro type of spoof, we choose 2 of their micro type of spoof for training, and the others for testing. Specifically, ``A4'', ``face mask'' and ``PC''  are selected for testing. \textbf{2)} \textit{Protocol 2 - } Protocol 2 evaluates the effect of input sensor variations. According to imaging quality, we split input sensors into three groups: \textit{low-quality sensor}, \textit{middle-quality sensor} and \textit{high-quality sensor}\footnote{Please refer to supplementary for the detailed input sensors information.}. Since we need to test on three different kinds of sensor and the average performance of FPR-Recall is hard to measure, we do not include FPR-Recall in the evaluation metrics of protocol 2. Table \ref{tab:protocol} shows the performance under each protocol.

\setlength{\tabcolsep}{5pt}
\begin{table}[t]
\centering
\caption{Cross-domain benchmark results of CelebA-Spoof. \textbf{Bolds} are the best results; $\uparrow$ means bigger value is better; $\downarrow$ means smaller value is better}
\label{tab:protocol}
\resizebox{\textwidth}{!}{%
\begin{tabular}{cccccccccc}
\toprule
\multirow{2}{*}{Protocol} &
  \multirow{2}{*}{Model} &
  \multicolumn{3}{c}{Recall (\%) $\uparrow$} &
  \multirow{2}{*}{AUC$\uparrow$} &
  \multirow{2}{*}{EER (\%)$\downarrow$} &
  \multirow{2}{*}{APCER (\%)$\downarrow$} &
  \multirow{2}{*}{BPCER (\%)$\downarrow$} &
  \multirow{2}{*}{ACER (\%)$\downarrow$} \\ \cmidrule{3-5}
 &
   &
  FPR = 1\% &
  FPR = 0.5\% &
  FPR = 0.1\% &
   &
   &
   &
   &
   \\ \midrule
   
    %  \multirow{3}{*}{\begin{tabular}[c]{@{}c@{}}Train on spoof type: \\ image, poster,upper body mask, pad,phone\\ Test on spoof type:A4,face mask,PC\end{tabular}} &
   
\multirow{3}{*}{1} &

  Baseline &
  93.7 &
  86.9 &
  69.6 &
  \textbf{0.996} &
  2.5 &
  5.7 &
  2.52 &
  4.11 \\ %\cline{3-11} 
   &
    AENet$_{\mathcal{C}, \mathcal{G}}$ &
  93.3 &
  88.6 &
  \textbf{74.0} &
  0.994 &
  2.5 &
  5.28 &
  2.41 &
  3.85 \\ %\cline{3-11} 
   &
  AENet$_{\mathcal{C}, \mathcal{S}}$ &
  93.4 &
  89.3 &
  71.3 &
  \textbf{0.996} &
  2.4 &
  5.63 &
  2.42 &
  4.04  \\ 
  &  AENet$_{\mathcal{C}, \mathcal{S}, \mathcal{G}}$ &
  \textbf{95.0} &
  \textbf{91.4} &
  73.6 &
  0.995 &
  \textbf{2.1} &
  \textbf{4.09} &
  \textbf{2.09} &
  \textbf{3.09}  \\ \midrule
  
    % \multirow{3}{*}{\begin{tabular}[c]{@{}c@{}}Train on 2 types of sensor\\ Test on 1 type of sensor\end{tabular}} &

\multirow{3}{*}{2} &
  Baseline &
  \# &
  \# &
  \# &
  0.998 $\pm $0.002 &
  1.5$\pm $0.8 &
  8.53$\pm $2.6 &
  1.56$\pm $0.81 &
  5.05$\pm $1.42 \\ %\cline{3-11} 
   &
AENet$_{\mathcal{C}, \mathcal{G}}$ &
  \# &
  \# &
  \# &
  0.995$\pm $0.003 &
  1.6$\pm $4.5 &
  8.95$\pm $1.07 &
  1.67$\pm $0.9 &
  5.31$\pm $0.95 \\ %\cline{3-11} 
   &
AENet$_{\mathcal{C}, \mathcal{S}}$ &
 
  \# &
  \# &
  \# &
  0.997$\pm $0.002 &
  1.2$\pm $0.7 &
  \textbf{4.01$\pm $2.9} &
  1.24$\pm $0.67 &
  3.96$\pm $1.79  \\ 
&  AENet$_{\mathcal{C}, \mathcal{S}, \mathcal{G}}$ &
  
  \# &
  \# &
  \# &
  \textbf{0.998$\pm $0.002} &
  \textbf{1.3$\pm $0.7} &
  4.94$\pm $3.42 &
  \textbf{1.24$\pm $0.73} &
  \textbf{3.09$\pm $2.08}  \\
  \bottomrule
\end{tabular}%
}
\end{table}

\subsection{Cross-Dataset Benchmark}
In this subsection, we perform cross-dataset testing on CelebA-Spoof and CASIA-MFSD dataset to further construct the cross-dataset benchmark. On the one hand, we offer a quantitative result to measure the quality of our dataset. On the other hand, we can evaluate the generalization ability of different methods according to this benchmark. The current largest face anti-spoofing dataset CASIA-SURF \cite{casiasurf} adopted \textit{FAS-TD-SF} \cite{depth:FAS-TD-SF} (which is trained on SiW or CASIA-SURF and tested on CASIA-MFSD) to demonstrate the quality of CASIA-SURF. Following this setting, we first train AENet$_{\mathcal{C}, \mathcal{G}}$, AENet$_{\mathcal{C}, \mathcal{S}}$ and AENet$_{\mathcal{C}, \mathcal{S}, \mathcal{G}}$ based on CelebA-Spoof and then test them on CASIA-MFSD to evaluate the quality of CelebA-Spoof. As shown in Table \ref{tab:cross testing}, we can conclude that: \textbf{1)} The diversity and large quantities of CelebA-Spoof drastically boosts the performance of vanilla model; a simple ResNet-18 achieves state-of-the-art cross-dataset performance. \textbf{2)} Comparing to geometric information, semantic information equips the model with better generalization ability.

\setlength{\tabcolsep}{5pt}
\begin{table}[t]
\centering
\caption{Cross-dataset benchmark results. AENet$_{\mathcal{C},\mathcal{S},\mathcal{G}}$ based on ResNet-18 achieves the best generalization performance. \textbf{Bolds} are the best results; $\uparrow$ means bigger value is better; $\downarrow$ means smaller value is better}
\label{tab:cross testing}
\centering
\resizebox{0.5\textwidth}{!}{%
\begin{tabular}{@{}cccc@{}}
\toprule
Model    &  Training     & Testing     & HTER (\%) $\downarrow$      \\ \midrule
FAS-TD-SF \cite{depth:FAS-TD-SF}  & SiW          & CASIA-MFSD & 39.4          \\ 
FAS-TD-SF \cite{depth:FAS-TD-SF}   & CASIA-SURF   & CASIA-MFSD & 37.3          \\ 
AENet$_{\mathcal{C},\mathcal{S}, \mathcal{G}}$   & SiW   & CASIA-MFSD & 27.6         \\ \midrule
Baseline&  CelebA-Spoof & CASIA-MFSD &14.3          \\ 
AENet$_{\mathcal{C},\mathcal{G}}$ &
 CelebA-Spoof & CASIA-MFSD & 14.1         \\ 
AENet$_{\mathcal{C},\mathcal{S}}$ &
 CelebA-Spoof & CASIA-MFSD & 12.1 \\ 
AENet$_{\mathcal{C},\mathcal{S}, \mathcal{G}}$ &
CelebA-Spoof & CASIA-MFSD & \textbf{11.9} \\\bottomrule 
\end{tabular}%
}
\end{table}

\section{Conclusion}
In this paper, we construct a large-scale face anti-spoofing dataset, \textbf{CelebA-Spoof}, with 625,537 images from 10,177 subjects, including 43 rich attributes on face, illumination, environment and spoof types. We believe CelebA-Spoof would be a significant contribution to the community of face anti-spoofing. Based on these rich attributes, we further propose a clean yet powerful multi-task framework, namely \textbf{AENet}. Through AENet, we conduct extensive experiments to explore the roles of semantic information and geometric information in face anti-spoofing. To support comprehensive evaluation and diagnosis, we establish three versatile benchmarks to evaluate the performance and generalization ability of various methods under different carefully-designed protocols. With several valuable observations revealed, we demonstrate the effectiveness of CelebA-Spoof and its rich attributes which can significantly facilitate future research.

\section*{Acknowledgments}
This work is supported in part by
SenseTime Group Limited, in part by National Science Foundation of China Grant No. U1934220 and 61790575, and the project “Safety data acquisition equipment for industrial enterprises No.134”. The corresponding author is Jing Shao.

\clearpage

\bibliographystyle{splncs04}
\bibliography{eccv2020submission.bbl}

\section{Appendix}
\subsection{Detail Information of CelebA-Spoof Dataset}
\label{App:dataset} 
\begin{figure}[h]
\centering
\includegraphics[width = \textwidth]{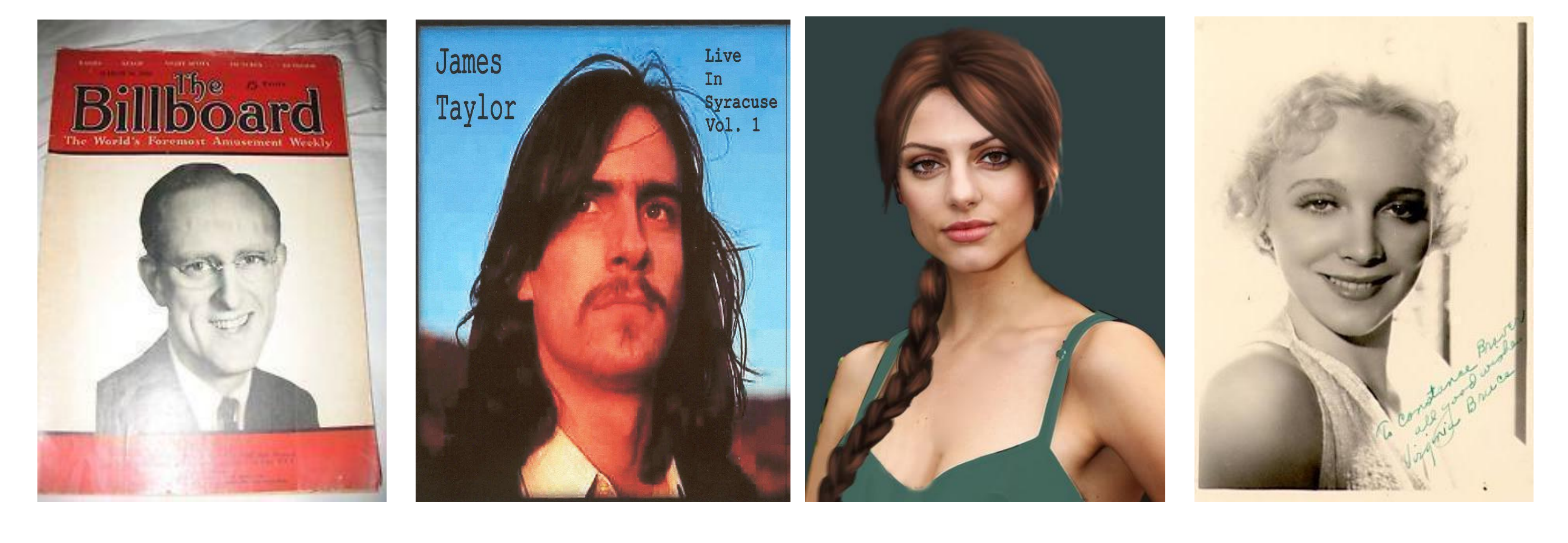}
\caption{Representative examples of the ``spoof '' images in CelebA}
\label{fig:spoof_image_in_CelebA}
\end{figure}

\noindent\textbf{Spoof Images in CelebA.}
As shown in Figure~\ref{fig:spoof_image_in_CelebA}. In CelebA~\cite{CelebA}, there are 347 ``spoof'' images, including poster, advertisements and portrait \textit{etc.} For spoof instruments selection and live data collection on CelebA-Spoof, we manually examine these images and remove them.

\setlength{\tabcolsep}{3pt}
\begin{table}[h]
\centering
\LARGE
\ra{1.3}
\caption{Input sensor split in CelebA-Spoof, there are 24 different input sensors which are split into 3 groups based on image quality} 
\label{tab:input sensor split}
\resizebox{\textwidth}{!}{%
\begin{tabular}{@{}c|cccc||c|cccc||c|cccc@{}}
\hline

 &
  Sensor &
  Dataset &
  Pix. (MP) &
  Release &
   &
  Sensor &
  Dataset &
  Pix. (MP) &
  Release &
   &
  Sensor &
  Dataset &
  Pix. (MP) &
  Release \\ \hline
\multirow{11}{*}{\begin{tabular}[c]{@{}c@{}}Low-Quality\\ Sensor\end{tabular}} &
  Honor V8 &
  train test val &
  1200 &
  2016 &
\multirow{11}{*}{\begin{tabular}[c]{@{}c@{}}Middle-Quality\\ Sensor\end{tabular}} &
  vivo X20 &
  train test val &
  1200 &
  2018 &
\multirow{11}{*}{\begin{tabular}[c]{@{}c@{}}High-Quality\\ Sensor\end{tabular}} &
  \multirow{4}{*}{HUAWEI P30} &
  \multirow{4}{*}{train test val} &
  \multirow{4}{*}{4000} &
  \multirow{4}{*}{2019} \\
 &
  OPPO R9 &
  train test val &
  1300 &
  2016 &
   &
  Gionee S11 &
  train test val &
  1300 &
  2018 &
   &
   &
   &
   &
   \\
 &
  HUAWEI MediaPad M5 &
  train test &
  1200 &
  2016 &
   &
  vivo Y85 &
  train val &
  1600 &
  2018 &
   &
   &
   &
   &
   \\
 &
  Xiaomi Mi Note3 &
  train test val &
  1200 &
  2016 &
   &
  Hisense H11 &
  train val &
  2000 &
  2018 &
   &
   &
   &
   &
   \\
 &
  Gionee S9 &
  train test val &
  1300 &
  2016 &
   &
  iphone XR &
  train &
  1200 &
  2018 &
   &
  \multirow{3}{*}{meizu 16S} &
  \multirow{3}{*}{train test val} &
  \multirow{3}{*}{4800} &
  \multirow{3}{*}{2019} \\
 &
  Logitech C670i &
  train &
  1200 &
  2016 &
   &
  \multirow{2}{*}{OPPO A5} &
  \multirow{2}{*}{train} &
  \multirow{2}{*}{1300} &
  \multirow{2}{*}{2018} &
   &
   &
   &
   &
   \\
 &
  ThinkPad T450 &
  train &
  800 &
  2016 &
   &
   &
   &
   &
   &
   &
   &
   &
   &
   \\
 &
  Moto X4 &
  train test val &
  1200 &
  2017 &
   &
  OPPO R17 &
  train &
  1600 &
  2018 &
   &
  \multirow{4}{*}{vivo NEX 3} &
  \multirow{4}{*}{train} &
  \multirow{4}{*}{6400} &
  \multirow{4}{*}{2019} \\
 &
  vivo X7 &
  train test val &
  1200 &
  2017 &
   &
  OPPO A3 &
  train test val &
  1200 &
  2019 &
   &
   &
   &
   &
   \\
 &
  Dell 5289 &
  train &
  800 &
  2017 &
   &
  Xiaomi 8 &
  train test val &
  1200 &
  2019 &
   &
   &
   &
   &
   \\
 &
  OPPO A73 &
  train &
  1600 &
  2017 &
   &
  vivo Y93 &
  train test val &
  1300 &
  2019 &
   &
   &
   &
   &
  \\ \hline
\end{tabular}%
}
\end{table}

\noindent\textbf{Input Sensor Split.} As shown in Table~\ref{tab:input sensor split}, according to imaging quality, we split 24 input sensors into 3 groups: \textit{low-quality sensor}, \textit{middle-quality sensor} and \textit{high-quality sensor}. In detail, an input sensor is not necessarily used in the all training, verification and testing set, so we specify which dataset these input sensors would cover. Specifically, for cross-domain benchmark in CelebA-Spoof, only input sensors which are both used in training set and testing set are selected.

\subsection{Experimental Details}
\noindent\textbf{Formulations of  Evaluation Metrics.}
To establish a comprehensive benchmark, we unify 7 commonly used metrics (\textit{i.e.}~APCER, BPCER, ACER, EER, HTER, AUC and FPR@Recall). Besides AUC, EER and FPR@Recall which are the most common metrics of classification tasks, we list definitions and formulations of other metrics. 1) \textit{APCER, BPCER and ACER.} Refer to~\cite{oulu-NPU,SiW-M}, Attack Presentation Classification Error Rate (APCER) is used to evaluate the classification performance of models for spoof images. Bona Fide Presentation Classification Error Rate (BPCER) is used to evaluate the classification performance of models for live images: 
\begin{equation}
APCER_{\mathcal{S}^\text{s}} = \frac{1}{N_{\mathcal{S}^\text{s}}}\sum_{i=1}^{N_{\mathcal{S}^\text{s}}}(1-Res_{i})
\end{equation}

% \begin{equation}
% APCER = \text{max}(APCER_{\mathcal{S}^{\text{s}_1}},APCER_{\mathcal{S}^{\text{s}_2}}...APCER_{\mathcal{S}^{\text{s}_k}})
% \end{equation}

\begin{equation}
APCER = \frac{1}{N_{spf.}}\sum_{i=1}^{N_{spf.}}(1-Res_{i})
\end{equation}

\begin{equation}
BPCER_{\mathcal{S}^\text{f}} = \frac{1}{N_{\mathcal{S}^\text{f}}}\sum_{i=1}^{N_{\mathcal{S}^\text{f}}}Res_{i}
\end{equation}

\begin{equation}
BPCER = \frac{1}{N_{liv.}}\sum_{i=1}^{N_{liv.}}Res_{i}
\end{equation}

\begin{equation}
ACER = \frac{(APCER+BPCER)}{2}
\end{equation}

where, $N_{\mathcal{S}^\text{s}}$ is the number of the spoof images of the given spoof type. $N_{\mathcal{S}^\text{f}}$ is the number of the live images of the given face attribute. $N_{liv.}$ is the number of all live images. $Res_{i}$ takes the value 1 if the ith images is classified as an spoof image and 0 if classified as live image. APCER$_{\mathcal{S}^\text{s}}$ is computed separately for each micro-defined spoof type (\textit{e.g.} ``photo'', ``A4'', ``poster''). 
Besides, in CelebA-Spoof, we define BPCER$_{\mathcal{S}^\text{f}}$ which is computed separately for each face attribute. To summarize the overall performance of live images and spoof images, the Average Classification Error Rate (ACER) is used, which is the average of the APCER and the BPCER at the decision threshold defined by the Equal Error Rate (EER) on the testing set. 2) \textit{HTER.} The aforementioned metrics are employed on intra-dataset (CelebA-Spoof) evaluation. For cross-dataset evaluation, HTER~\cite{LBP_TOP} is used extensively:
\begin{equation}
    HTER(D_{2}) = \frac{FAR(\tau (D1),(D_{2})) + FRR(\tau (D1),(D_{2}))}{2}
\end{equation}
where $\tau (D_{n})$ is a threshold, $D_n$ is the dataset, False Acceptance Rate (FAR) and False Rejection Rate (FRR) is the value in $D_2$. In cross-dataset evaluation, the value of $\tau (D_{n})$ is estimated on the EER using the testing set of the dataset $D_1$. In this equation, when $D_1 \neq D_2$, we have the cross-dataset evaluation. 

\begin{figure}[t]
\centering
\includegraphics[width = 0.8 \textwidth]{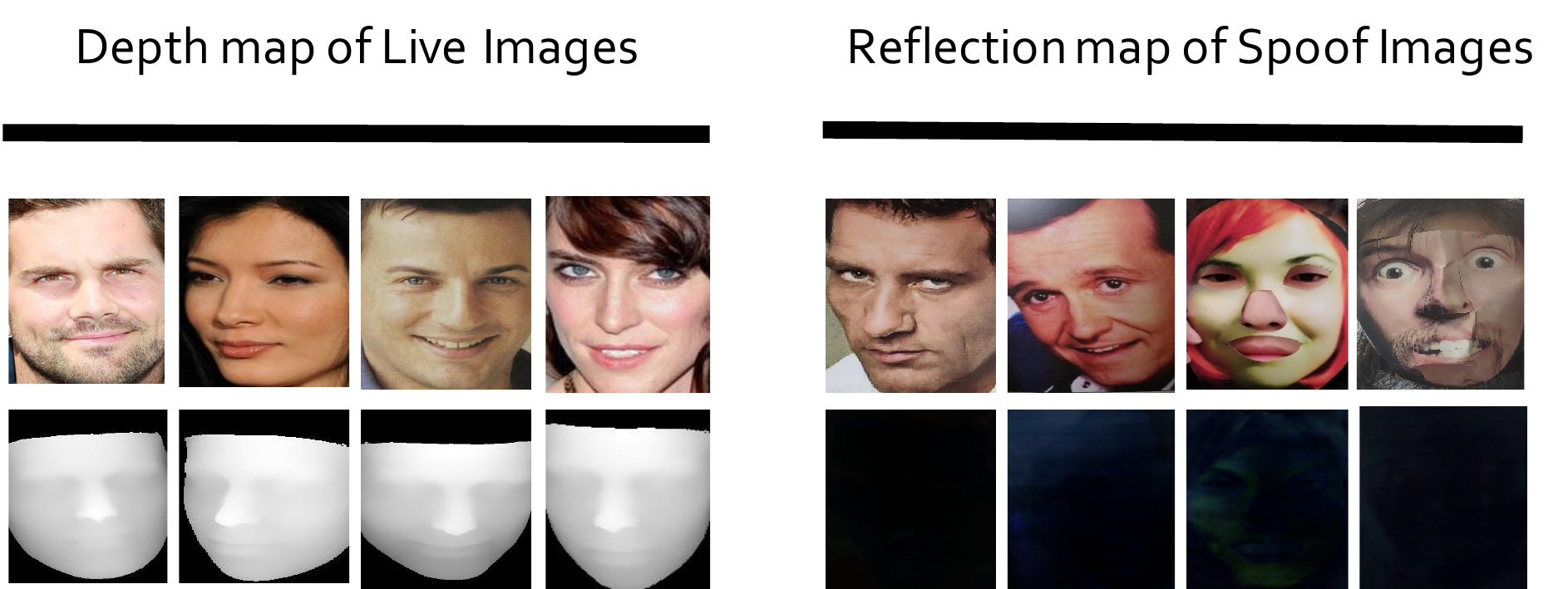}
\caption{All live images have depth maps, but only the second and the third spoof image has reflection artifacts. Zoom in for better visualization}
\label{fig:depth_reflet_GT}
\end{figure}

\noindent\textbf{The Limitations of Reflection Map.} For ablation study of geometric information, we do not use reflection maps as unique binary supervision. This is because only parts of spoof images show reflect artifacts as shown in Figure~\ref{fig:depth_reflet_GT}. In this figure, only the second and the third spoof image shows reflect artifacts, the reflection map for other spoof images is zero. However, each live image has its corresponding depth map.

\begin{table}[t]
\centering
\caption{The mAP result of single-task and multi-task. There is huge space to improve the learning of $\mathcal{S}^\text{i}$ in multi-task fashion. \textbf{Bolds} are the best results}
\label{tab:mAP}
\resizebox{0.3\textwidth}{!}{%
\begin{tabular}{@{}ccc@{}}
\toprule
Attribute                                       & Model   & mAP (\%)       \\ \midrule
\multirow{2}{*}{$\mathcal{S}^\text{s}$} & AENet$_{\mathcal{S}^\text{s}}$ & 45.7          \\  
                                                & AENet$_{\mathcal{C},\mathcal{S}}$ & \textbf{46.2} \\ \midrule
\multirow{2}{*}{$\mathcal{S}^\text{f}$}               & AENet$_{\mathcal{S}^\text{f}}$ & 68.5          \\ 
                                                & AENet$_{\mathcal{C},\mathcal{S}}$ & \textbf{70.5} \\ \midrule
\multirow{2}{*}{$\mathcal{S}^\text{i}$}         & AENet$_{\mathcal{S}^\text{i}}$ & \textbf{57.1} \\ 
                                                & AENet$_{\mathcal{C},\mathcal{S}}$ & 43.3          \\ \bottomrule
\end{tabular}%
}
\end{table}

% \begin{table}[htbp]
% \centering
% \ra{1.3}
% \caption{Detail information of data split of CelebA-Spoof}
% \label{tab:Datasplit}
% \resizebox{0.4\textwidth}{!}{
% \begin{tabular}{ccccc}
% \toprule
% \#              & Training & Validation & Testing & Total  \\ \midrule
%     Subject.    & 8192     & 985        & 1000    & 10177  \\ 
%     Live images. & 162462   & 19867      & 19923   & 202252 \\ 
%     Spoof images.  & 331943   & 44095      & 47247   & 423285 \\ 
%     All images.  & 494405   & 63962      & 67170   & 625537 \\ \bottomrule
%     \end{tabular}
%     }
%     \end{table}

% \setlength{\tabcolsep}{5pt}
% \begin{table}[t]
% \centering
% \ra{1.3}
% \caption{\footnotesize{Model setting in single-task and multi-task comparing}}
% \label{tab:table multi task and single task}
% \resizebox{0.6\textwidth}{!}{%
% \begin{tabular}{@{}ccccc@{}}
% \hline
%                                 & AENet$_{\mathcal{S}^\text{f}}$ &
%                          AENet$_{\mathcal{S}^\text{s}}$ &
%                          AENet$_{\mathcal{S}^\text{i}}$ & AENet$_{\mathcal{C},\mathcal{S}}$\\ \hline
% Live/Spoof              & &  &    &  $\surd$   \\
% Face Attribute                &   $\surd$  &  & & $\surd$  \\
% Spoof Type &  & $\surd$ &  & $\surd$ \\
% Illumination Conditions        &         &         & $\surd$   & $\surd$   \\ 
% \hline
% \end{tabular}%
% }
% \end{table}

\noindent\textbf{Multi Task and Single Task.} Besides ablation study of semantic information. We compare AENet$_{\mathcal{S}^\text{f}}$, AENet$_{\mathcal{S}^\text{s}}$ and AENet$_{\mathcal{S}^\text{i}}$ with AENet$_{\mathcal{C},\mathcal{S}}$ to explore whether multi-task learning can promote classification performance of these semantic information. In detail, as mentioned in model setting of Sec. \textit{Ablation Study on CelebA-Spoof}. AENet$_{\mathcal{S}^\text{f}}$, AENet$_{\mathcal{S}^\text{s}}$ and AENet$_{\mathcal{S}^\text{i}}$ are trained for classification of each semantic information.  As shown in the Table \ref{tab:mAP}. It shows that the mAP performance of $\mathcal{S}^\text{f}$ and the $\mathcal{S}^\text{s}$ in AENet$_{\mathcal{C},\mathcal{S}}$ is better than AENet$_{\mathcal{S}^\text{f}}$ and  AENet$_{\mathcal{S}^\text{s}}$. Specifically, these two semantic information are proven crucial to improve classification of live/spoof images in ablation study. Besides, the mAP performance of $\mathcal{S}^\text{i}$ of AENet$_{\mathcal{C},\mathcal{S}}$ is worse than AENet$_{\mathcal{S}^\text{i}}$. This is because we set the $\lambda=0.01$ of $\mathcal{S}^\text{i}$ in the multi-task training but $\lambda=1$ for all single task model. This small value let $\mathcal{S}^\text{i}$ difficult to converge in multi task learning. 

\subsection{Benchmark on Heavier Model}
\label{App:Benchmark}
In order to build a comprehensive benchmark, besides ResNet-18~\cite{resnet}, we also provide the corresponding results based on a heavier backbone, \textit{i.e.}~Xception~\cite{chollet2017xception}. All the results on the following 3 benchmarks are based on Xception. Detail information about benchmark based on ResNet-18 is shown in paper. 1) \textit{Intra-Dataset Benchmark.}~As shown in Table~\ref{tab:xception_benchmark1}, AENet$_{\mathcal{C},\mathcal{S},\mathcal{G}}$ based on Xception achieve better performance comparing to AENet$_{\mathcal{C},\mathcal{S},\mathcal{G}}$ based on ResNet-18, especially when FPR is smaller (\textit{i.e.}~FPR=0.5\% and FPR=0.1\%). This is because model with heavier parameters can achieve better robustness. 2) \textit{Cross-domain Benchmark.}~As shown in  Table~\ref{tab:xception_benchmark2}, AENet$_{\mathcal{C},\mathcal{S},\mathcal{G}}$ based on Xception achieve the better performance than AENet$_{\mathcal{C},\mathcal{S},\mathcal{G}}$ based on ResNet-18. And in protocol 1, comparing to baseline based on Xception.  AENet$_{\mathcal{C},\mathcal{S},\mathcal{G}}$ based on Xception outperforms baseline by 67.3\% in APCER. 3) \textit{Cross-dataset Benchmark.}~As shown in Table~\ref{tab:xception_benchmark3}. Performance of models based on Xception is worse than models based on ResNet-18. This is because models with heavier parameters tend to fit the training data.

\setlength{\tabcolsep}{5pt}
\begin{table}[t]
\centering
\ra{1.3}
\caption{Intro-dataset Benchmark results of CelebA-Spoof. AENet$_{\mathcal{C},\mathcal{S},\mathcal{G}}$ achieved the best result. \textbf{Bolds} are the best results; $\uparrow$ means bigger value is better; $\downarrow$ means smaller value is better} 
\label{tab:xception_benchmark1}
\resizebox{\textwidth}{!}{%
\begin{tabular}{@{}cccccccccc@{}}
\toprule 
\multirow{2}{*}{Model} &
\multirow{2}{*}{Parm. (MB)} &
  \multicolumn{3}{c}{Recall (\%)$\uparrow$} &
  \multirow{2}{*}{AUC$\uparrow$} &
  \multirow{2}{*}{EER (\%)$\downarrow$} &
  \multirow{2}{*}{APCER (\%)$\downarrow$} &
  \multirow{2}{*}{BPCER (\%)$\downarrow$} &
  \multirow{2}{*}{ACER (\%)$\downarrow$} \\ \cmidrule{3-5}
         &    & FPR = 1\% & FPR = 0.5\%   & FPR = 0.1\%   &        &       &      &      &      \\ \midrule
AENet$_{\mathcal{C},\mathcal{G}}$    & 79.9     & 
98.3      & 97.2          & 91.4          & \textbf{0.9982} & 1.2 & 4.98 & 1.26 & 3.12 \\ %\hline
AENet$_{\mathcal{C},\mathcal{S}}$     & 79.9   & 
98.5      & 97.8 & \textbf{94.3} & 0.9980 & 1.3 & 4.22 & 1.21 & 2.71 \\ 
AENet$_{\mathcal{C},\mathcal{S},\mathcal{G}}$     & 79.9   & \textbf{99.2}     & \textbf{98.4} & 94.2 & 0.9981 & \textbf{0.9} & \textbf{3.72} & \textbf{0.82} & \textbf{2.27}\\ \bottomrule %\hline
\end{tabular}%
}
\end{table}

\setlength{\tabcolsep}{5pt}
\begin{table}[t]
\centering
\ra{1.3}
\caption{Cross-domain benchmark results of CelebA-Spoof. \textbf{Bolds} are the best results; $\uparrow$ means bigger value is better; $\downarrow$ means smaller value is better}
\label{tab:xception_benchmark2}
\resizebox{\textwidth}{!}{%
\begin{tabular}{cccccccccccc}
\toprule
\multirow{2}{*}{Protocol} &
  \multirow{2}{*}{Model} &
  \multicolumn{3}{c}{Recall (\%) $\uparrow$} &
  \multirow{2}{*}{AUC$\uparrow$} &
  \multirow{2}{*}{EER (\%)$\downarrow$} &
  \multirow{2}{*}{APCER (\%)$\downarrow$} &
  \multirow{2}{*}{BPCER (\%)$\downarrow$} &
  \multirow{2}{*}{ACER (\%)$\downarrow$} \\ \cmidrule{3-5}
   &
   &
  FPR = 1\% &
  FPR = 0.5\% &
  FPR = 0.1\% &
   &
   &
   &
   &
   \\ \midrule
\multirow{3}{*}{1} &
  Baseline &
  94.6 &
  92.3 &
  \textbf{86.4} &
  0.985 &
  3.8 &
  9.19 &
  3.84 &
  6.515 \\ %\cline{3-11} 
 &
    AENet$_{\mathcal{C},\mathcal{G}}$ &
  93.7 &
  89.7 &
  73.1 &
  0.984 &
  3.4 &
  7.66 &
  3.11 &
  5.39 \\ %\cline{3-11} 
 &
    AENet$_{\mathcal{C},\mathcal{S}}$ &
  96.5 &
  \textbf{93.1} &
  83.4 &
  0.992 &
  2.3 &
  3.78 &
  1.8 &
  2.79 \\ 
   &
AENet$_{\mathcal{C},\mathcal{S},\mathcal{G}}$ &
  \textbf{96.9} &
  93.0 &
  83.5 &
  \textbf{0.996} &
  \textbf{1.8} &
  \textbf{3.00} &
  \textbf{1.48} &
  \textbf{2.24} \\ \midrule
\multirow{3}{*}{2} &
  Baseline &
  \# &
  \# &
  \# &
  0.996$\pm $0.003 &
  1.8$\pm $0.9 &
  7.44$\pm $2.62 &
  1.81$\pm $0.9 &
  4.63$\pm $1.66 \\ %\cline{3-11} 
 &
AENet$_{\mathcal{C},\mathcal{G}}$ &
  \# &
  \# &
  \# &
  0.994$\pm $0.006 &
  1.7$\pm $0.6 &
  9.16$\pm $1.97 &
  1.56$\pm $1.68 &
  5.36$\pm $1.23 \\ %\cline{3-11} 
 &

AENet$_{\mathcal{C},\mathcal{S}}$ &
  \# &
  \# &
  \# &
  0.996$\pm $0.003 &
  1.2$\pm $0.9 &
  5.08$\pm $4.41 &
  \textbf{0.95$\pm $0.68} &
  4.02$\pm $2.6 \\ 
  &
  AENet$_{\mathcal{C},\mathcal{S},\mathcal{G}}$ &
  \# &
  \# &
  \# &
  \textbf{0.997$\pm $0.003} &
  \textbf{1.3$\pm $1.2} &
  \textbf{4.77$\pm $4.12} &
  1.23$\pm $1.06 &
  \textbf{3.00$\pm $2.9} \\  \bottomrule
\end{tabular}%
}
\end{table}

\setlength{\tabcolsep}{5pt}
\begin{table}[!t]
\centering
\ra{1.3}
\caption{Cross-dataset benchmark results of CelebA-Spoof. AENet$_{\mathcal{C},\mathcal{S},\mathcal{G}}$  achieves the best generalization performance. \textbf{Bolds} are the best results; $\uparrow$ means bigger value is better; $\downarrow$ means smaller value is better}
\label{tab:xception_benchmark3}
\centering
\resizebox{0.4\textwidth}{!}{%
\begin{tabular}{@{}cccc@{}}
\toprule
Model      & Training     & Testing     & HTER (\%) $\downarrow$      \\ \midrule
Baseline & CelebA-Spoof & CASIA-MFSD & 20.1          \\ 
AENet$_{\mathcal{C},\mathcal{G}}$  & CelebA-Spoof & CASIA-MFSD & 18.2          \\ 
AENet$_{\mathcal{C},\mathcal{S}}$  &  CelebA-Spoof & CASIA-MFSD & 17.7 \\ 
AENet$_{\mathcal{C},\mathcal{S},\mathcal{G}}$  &  CelebA-Spoof & CASIA-MFSD & \textbf{13.1} \\ \bottomrule 
\end{tabular}%
}
\end{table}

\end{document}